%% file: cr.tex
\documentclass[10pt,twocolumn,letterpaper]{article}

\usepackage{iccv}      

\input{preamble}

\definecolor{iccvblue}{rgb}{0.21,0.49,0.74}
\usepackage[pagebackref,breaklinks,colorlinks,allcolors=iccvblue]{hyperref}
\usepackage{multirow}
\usepackage{multicol}
\usepackage{threeparttable}
\usepackage{tabularx}
\usepackage{booktabs}
\usepackage{makecell}
\usepackage{colortbl}
\usepackage{graphicx}
\usepackage{amsmath}
\usepackage{bbding}
\usepackage{listings}
\usepackage{arydshln}
\usepackage[ruled,linesnumbered]{algorithm2e}
\usepackage{color}

\def\paperID{7329} 
\newcommand{\figref}[1]{Fig.~\ref{#1}}
\newcommand{\tabref}[1]{Table~\ref{#1}}
\newcommand{\eqnref}[1]{Eq.~(\ref{#1})}

\def\confName{ICCV}
\def\confYear{2025}
\def\modelName{TopoTTA}
\def\stageone{Topological Structure Adaptation}
\def\stagetwo{Topological Continuity Refinement}
\title{TopoTTA: Topology-Enhanced Test-Time Adaptation for \\ Tubular Structure Segmentation}

\author{Jiale Zhou$^{1,2}$\!, Wenhan Wang$^{3}$\!, Shikun Li$^{2}$\!, Xiaolei Qu$^{3}$\!, Xin Guo$^{3}$\!, Yizhong Liu$^{3}$\!, \\Wenzhong Tang$^{3}$\!, Xun Lin$^{2,3,}$\thanks{Corresponding authors.}\,\,,
Yefeng Zheng$^{2,}$\footnotemark[1]
\\
$^1$Zhejiang University ~\quad
$^2$Westlake University ~\quad
$^3$Beihang University \\
{\tt\small \{zhoujiale, zhengyefeng\}@westlake.edu.cn ~\quad linxun@buaa.edu.cn}
}

\begin{document}
\maketitle

\begin{abstract}

Tubular structure segmentation (TSS) is important for various applications, such as hemodynamic analysis and route navigation. 
Despite significant progress in TSS, domain shifts remain a major challenge, leading to performance degradation in unseen target domains. Unlike other segmentation tasks, TSS is more sensitive to domain shifts, as changes in topological structures can compromise segmentation integrity, and variations in local features distinguishing foreground from background (e.g., texture and contrast) may further disrupt topological continuity. To address these challenges, we propose \textbf{Topo}logy-enhanced \textbf{T}est-\textbf{T}ime \textbf{A}daptation (\textbf{TopoTTA}), the first test-time adaptation framework designed specifically for TSS. TopoTTA consists of two stages: \textit{Stage~1} adapts models to cross-domain topological discrepancies using the proposed 
\textbf{Topo}logical \textbf{M}eta \textbf{D}ifference \textbf{C}onvolutions (\textbf{TopoMDC}s), which enhance topological representation without altering pre-trained parameters; Stage~2 improves topological continuity by a novel \textbf{Topo}logy \textbf{H}ard sample \textbf{G}eneration (\textbf{TopoHG}) strategy and prediction alignment on hard samples with pseudo-labels in the generated \textit{pseudo-breaks}. Extensive experiments across four scenarios and ten datasets demonstrate TopoTTA's effectiveness in handling topological distribution shifts, achieving an average improvement of 31.81\% in clDice. TopoTTA also serves as a plug-and-play TTA solution for CNN-based TSS models.

\end{abstract}

\section{Introduction}
\label{introduction}
Tubular structure segmentation (TSS), also known as curvilinear object segmentation ~\cite{bibiloni2016survey}, plays a critical role in various applications (e.g., hemodynamic analysis~\cite{cs2net, tan2022retinal} and route planning~\cite{road1, road2}). Although many TSS methods have been proposed and achieved inspiring performance~\cite{tss_eccv, topology_uc}, domain shifts remain a common challenge in real-world scenarios~\cite{ds3} due to discrepancies in imaging devices~\cite{ds1, ds2} and sample heterogeneity~\cite{Heterogeneous}.

\begin{figure}[t]
    \centering
    \includegraphics[width=0.9 \linewidth]{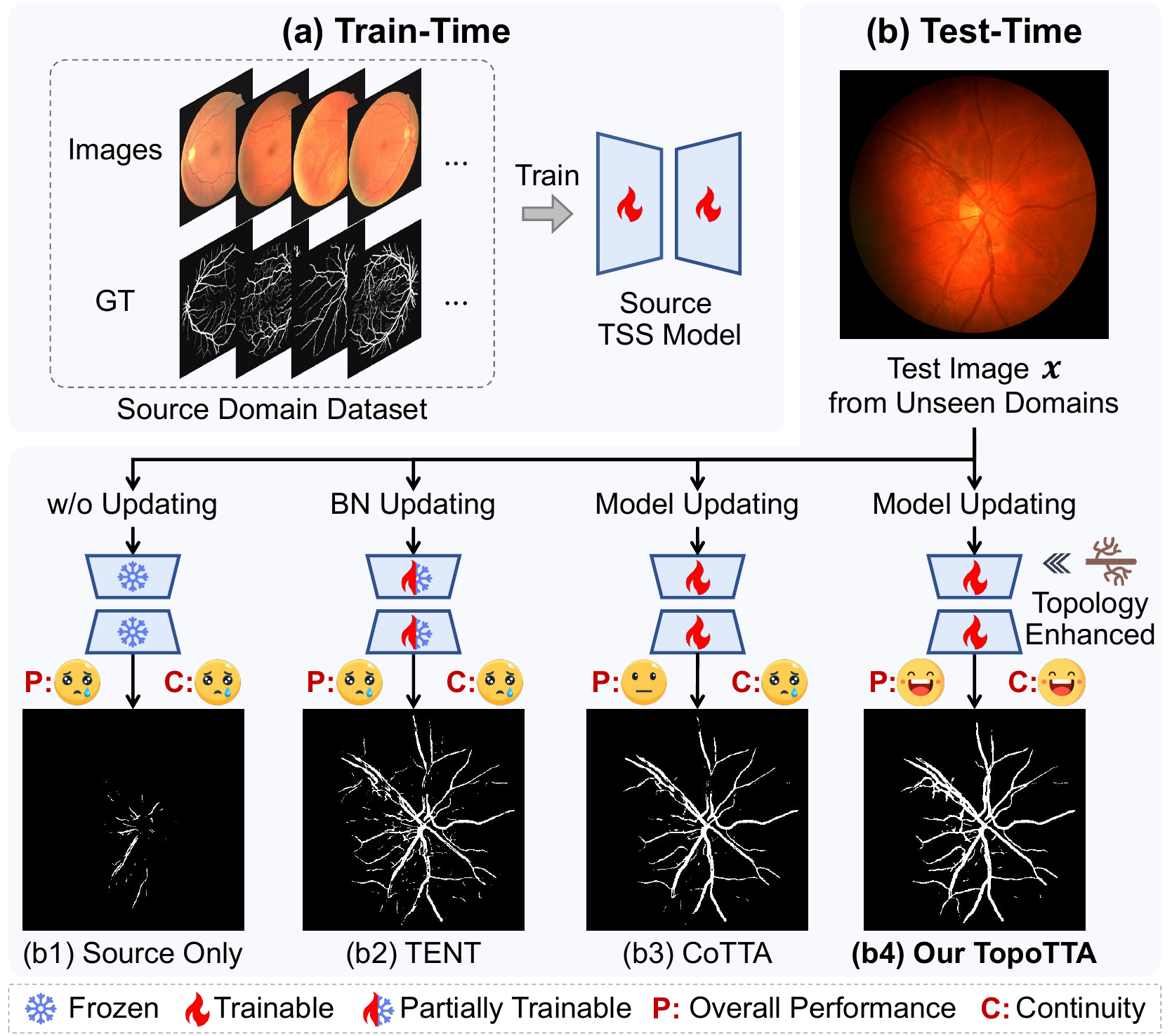}
    \vspace*{-2mm}
    \caption{Comparison between TopoTTA and existing TTA solutions. (a) represents the model during the source domain training phase, while (b) denotes the model in the target domain testing phase. (b1-b4) show the results of directly testing the source-trained model, the TENT~\cite{tta_tent} method with updated BN layers, the CoTTA~\cite{tta_cotta} method using a teacher-student scheme, and our topology-enhanced approach, respectively.}
    \vspace*{-5mm}
    \label{fig:overview}
\end{figure}

Domain shifts undermine the effectiveness of TSS models trained on the source domain when applied to an unseen target domain~\cite{Retinal_dg, liot}. 
Unlike segmentation models trained for other tasks, e.g., semantic~\cite{liu2021semantic, yang2020semantic} or tumor segmentation~\cite{dong2020tumor,hidemia,zhou2023mp,umed2024,ehformer2024}, TSS models are more sensitive to domain shifts~\cite{liot}. 
Beyond the common variations such as contrast and noise distribution~\cite{tss_noise,umb2025}, TSS faces additional challenges related to unseen topological structural characteristics, including variations in trajectory, curvature, branching patterns, and thickness~\cite{qi2023dscnet, tss_problem}. 
These factors degrade segmentation performance and even disrupt the topological continuity of the structures~\cite{deepclose} (see Fig.~\ref{fig:overview}~(b1)), which may mislead critical downstream applications~\cite{tan2022retinal, road1, rose}.

Recently, a special domain adaptation (DA) paradigm, namely test-time adaptation (TTA)~\cite{tta_survey, tta_SAR, tta_tipi}, has been proposed and successfully applied to various image segmentation tasks, including semantic~\cite{tta_cotta, tta_diga, tta_sita} and medical image segmentation~\cite{tta_vptta, tta_testfit}. 
Unlike other DA paradigms, TTA performs continuous online adaptation during the test stage (after model deployment)~\cite{tta_rotta}. 
It uses extremely small amounts of data (e.g., a single sample or batch), aligning better with real-world scenarios where data must be processed sequentially~\cite{tta_tent}.

Most TTA methods update pre-trained models using self-supervised methods~\cite{tta_survey}, which can typically be categorized into normalization-based methods~\cite{tta_tent, tta_domainadaptor} and teacher-student schemes~\cite{tta_rmt, tta_cotta}.
Normalization-based methods update normalization layer statistics~\cite{tta_ecotta} (e.g., batch/layer normalization (BN/LN)) or leverage the difference in normalization statistics before and after adaptation to update external model parameters~\cite{tta_vptta, tta_prompt_align}.
These methods are initially designed for classification tasks, and their performance may degrade when applied to segmentation tasks with severe pixel-level class imbalance~\cite{tta_imbalance}.
The second category, teacher-student schemes (based on pseudo-labels), updates all model parameters and achieves performance gains in general segmentation tasks~\cite{mm_question}. 
However, as shown in Fig.~\ref{fig:overview}, these methods perform poorly in TSS due to two unique challenges in cross-domain scenarios:
\begin{itemize}

\item \textbf{Challenge 1: Discrepant topological structures.} 
Tubular structures in the source and target domains can exhibit substantial topological differences (e.g., thickness, curvature, and branching patterns). 
Prior TTA methods' {general-purpose} model adaptation strategies struggle to effectively capture these cross-domain topological variations when adapting to individual samples.

\item \textbf{Challenge 2: Fragile topological continuity.} 
In cross-domain settings, the local characteristics (e.g., color, texture, and contrast) distinguishing foreground from background may change significantly~\cite{liot}. 
Existing TTA methods {lack targeted adaptation} to these local variations, resulting in catastrophic disruptions to topological continuity (see~\figref{fig:overview}) by misclassifying foreground pixels.

\end{itemize}

To address the aforementioned challenges, we propose \textbf{Topo}logy-Enhanced \textbf{T}est-\textbf{T}ime \textbf{A}daptation (\textbf{TopoTTA}) in this paper. It  includes two adaptation stages: topological structure adaptation~(\textit{Stage~1}) and topological continuity refinement~(\textit{Stage~2}).

\textit{Stage~1} aims to enhance the model's capability to adapt to discrepant topological structures by utilizing a topology-perceiving model adaptation strategy.
Inspired by the Central Difference Convolution (CDC)~\cite{cdc}, which captures differences between the central pixel and its neighbors to enhance local representations, we extend this concept to accommodate the elongated and diverse trajectories of tubular structures.
Specifically, we introduce \textbf{Topo}logical \textbf{M}eta \textbf{D}ifference \textbf{C}onvolutions (\textbf{TopoMDC}s), which perform a dual-pixel difference operation by extending the central pixel's interaction across eight directional neighbors. As shown in~\figref{fig:main}, we learn weighted combinations of different directional TopoMDCs (instead of the vanilla convolution) for different regions to enhance the representations of hard-to-detect tubular branches, improving adaptability to discrepant topological structures.

Within \textit{Stage~2}, we design a \textbf{Topo}logy \textbf{H}ard sample \textbf{G}eneration (\textbf{TopoHG}) strategy to construct samples with \textit{pseudo-breaks}.
These \textit{pseudo-breaks} mislead the models to make weakened predictions with poor topological continuity.
We then emphasize the local consistency between the predictions and original pseudo-labels in these hard regions.
This stage encourages the models to adapt the characteristics distinguishing foreground from background and helps address continuity destructions in the target domain.

Our contributions are as follows:
\begin{itemize}
\item We propose the first TTA framework for tubular structure segmentation, named TopoTTA, which effectively adapts to unseen domains by enhancing topological information. 
\item We design eight directional \textbf{Topo}logical \textbf{M}eta \textbf{D}ifference \textbf{C}onvolutions (\textbf{TopoMDC}s) within TopoTTA, and learn how to adaptively combine different TopoMDCs based on the topological features of the test sample.
\textbf{TopoMDC}s enhance adaptability to unseen topological structures.
\item We propose a \textbf{Topo}logy \textbf{H}ard sample \textbf{G}eneration strategy for TopoTTA, namely \textbf{TopoHG}, which generates \textit{pseudo-breaks} at confidently predicted regions. We then align the predictions around these \textit{pseudo breaks} to the original pseudo labels, improving the topological continuity in the unseen domain.

\item Extensive experiments on four scenarios across ten datasets validate the effectiveness of TopoTTA in addressing topological distribution shifts, achieving an average performance improvement of 31.81\% in clDice. 
\end{itemize}

\section{Related Works}
\subsection{Tubular Structure Segmentation}
Tubular structures, characterized by their thin and long topological structures as well as complex layouts, usually cannot be accurately and continuously localized by general segmentation methods~\cite{pointscatter}.
To tackle these challenges, researchers have incorporated topological priors into model designs, including specialized deformable convolution kernels~\cite{qi2023dscnet}, attention mechanisms~\cite{miccai21topoattn,morformer}, and tailored model architectures~\cite{iccv21crackformer}.
Complementary to these module innovations, PointScatter~\cite{pointscatter} was introduced to represent tubular structures as point sets, addressing the flexibility limitations of masks constrained by fixed grids.
DconnNet~\cite{dconnnet} employed a unique design to separate directional subspaces from the latent space, enhancing feature representation for connectivity. 
Meanwhile, a growing body of works focused on novel loss functions~\cite{cldice, hu2022dmt, wang2019tubularloss, eccv24skeleton, cvpr24ciloss, nips22warping, iclr21dmt} that explicitly optimize the topological continuity of segmentation results. 
More recently, some research has started to explore the use of synthetic data~\cite{iccv23freecos, eccv24costg, mia23yolo} and the integration of additional topological information~\cite{tss_noise,tss_eccv} to further improve performance.
However, these TSS methods ignore the existence of domain shifts, resulting in suboptimal performance when applied to unseen domains.

\subsection{Test-Time Adaptation}
TTA adaptively adjusts the model at test time based on a small number of unlabeled samples from a different distribution. 
Early TTA methods primarily modified model representations by adjusting BN's statistics~\cite{tta_delta, tta_dua, tta_medbn, tta_diga}. 
\citet{tta_dua} adapted target domain features by updating BN statistics continuously. 
Most TTA methods used self-supervised learning to adjust pre-trained parameters, including normalization-based methods~\cite{tta_tent, tta_domainadaptor, tta_ecotta, tta_SAR, tta_vptta} and teacher-student (TS) schemes~\cite{tta_cotta, tta_rmt, tta_tesla, icra_zsh}. 
\citet{tta_tent} proposed TENT, an entropy minimization method that updates BN affine transformation parameters to improve confidence in the target domain. 
Motivated by TENT, some recent works achieved better performance~\cite{tta_domainadaptor,tta_vptta} by designing various strategies to fine-tune BN. 
\citet{tta_domainadaptor} incorporated source domain statistics and extended the entropy minimization loss to utilize target domain information better. 
More recently, \citet{tta_vptta} reduced generalization errors by aligning source and target statistics to update external model parameters. 
However, many TTA methods are designed for classification tasks, ignoring pixel-wise class imbalance. 
In segmentation tasks, where sample imbalance is more severe, simple statistical adjustments and minor parameter updates are easily dominated by the majority of background pixels, reducing segmentation performance~\cite{mm_question}. 
The TS scheme, which applies consistency regularization to update all model parameters, has shown effectiveness in general segmentation tasks. 
CoTTA~\cite{tta_cotta} used weight-averaged and augmentation-averaged predictions, further reducing error accumulation in continuous inference. 
However, these methods lack a focus on tubular structure and topological continuity, leading to performance decline and continuity destruction in TSS.

\begin{figure*}[t]
	\setlength{\abovecaptionskip}{0pt}
	\setlength{\belowcaptionskip}{-8.5pt}
	\centering 
	\includegraphics[width=1.0\textwidth]{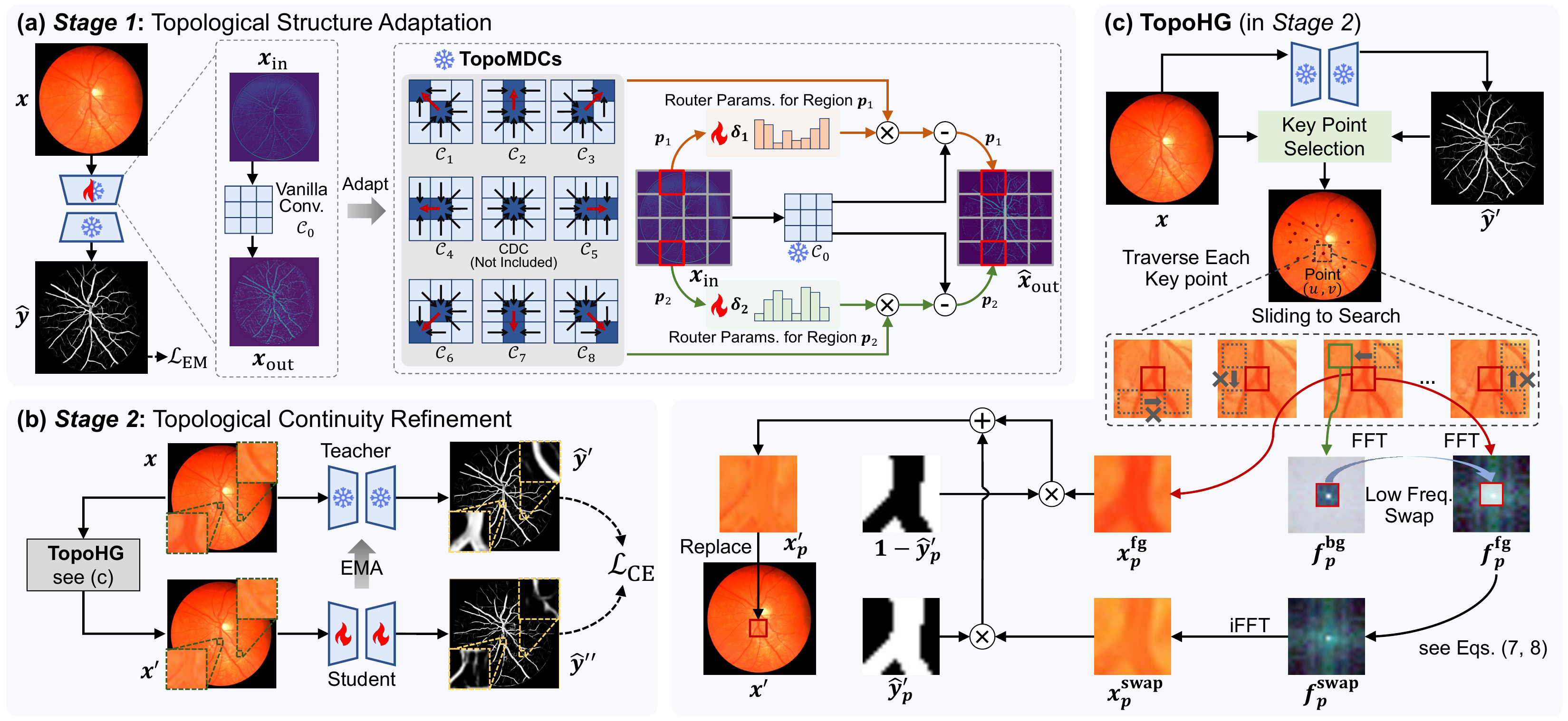}
	\caption{Overview of the proposed TopoTTA. TopoTTA consists of two stages: (a) \textit{Stage~1}: Topology structure adaptation with Topological Meta Difference Convolutions (TopoMDCs), and (b) \textit{Stage~2}: Topological continuity refinement. (c) Topology Hard sample Generation (TopoHG) constructs challenging samples with local \textit{pseudo-breaks}. CDC is short for central difference convolution.}
	\label{fig:main} 
    \vspace{-7pt}
\end{figure*}

\section{Proposed \modelName}

\subsection{Preliminary}

Let $\mathcal{D}^{s}=\{(\boldsymbol x_i^{s}, \boldsymbol y_i^{s})\}_{i=1}^{N^{s}}$ denote the labeled source domain dataset, where $\boldsymbol x_i^{s} \in \mathbb R^{H\times W \times C}$ is an image with height $H$, width $W$, and channel number $C$, and $\boldsymbol y_i^{s} \in \{0,1\}^{H\times W \times C}$ denotes the corresponding ground-truth label. 
Given a TSS model $\mathcal F(\cdot; \theta^s)$ trained on $\mathcal{D}^{s}$:
\begin{equation}
    \small
    \theta^s = \mathop{\arg\min}\limits_{\theta^s}\mathbb{E}_{(\boldsymbol{x}^s, \boldsymbol{y}^s)\sim \mathcal D^{s}}\Big[\mathcal L_{seg}(\mathcal F(\boldsymbol{x}^s; \theta^s), \boldsymbol{y}^s)\Big],
\end{equation}
where $\mathcal L_{seg}(\cdot, \cdot)$ can be any TSS loss function.
Let $\mathcal{D}^{t}=\{\boldsymbol x_i^{t}\}_{i=1}^{N^{t}}$ represent the unlabeled target domain dataset. 
During testing, TTA methods aim to update the model parameters for each input image $\boldsymbol x_i^t \in \mathcal{D}^{t}$ and produce segmentation outputs $\boldsymbol{\hat{y}}_i^t = \mathcal F(\boldsymbol{x}^t_i; \theta^t_{i-1})$ as follows:
\begin{equation}
    \small
    \theta^{t}_i = \mathop{\arg\min}\limits_{\theta^t_{i-1}} \mathcal L_{tta}(\mathcal F(\cdot; \theta^t_{i-1}), \boldsymbol x_i^t ),
\end{equation}
where $\theta^{t}_i$ are the parameters of the TSS model $\mathcal F(\cdot;\cdot)$ at the $i$-th iteration, and  $\mathcal L_{tta}(\cdot,\cdot)$ denotes the self-supervised loss function for TTA. Before performing TTA, we set $\theta_0^t = \theta^s$.
 
To improve readability, we use the simplified notation $\boldsymbol{x}$ instead of $\boldsymbol{x}^t_i$ in the following sections and figures.

\subsection{Overview of \modelName}

As discussed in Section~\ref{introduction}, existing TTA methods cannot effectively segment tubular structures due to two major challenges: discrepant topological structures and fragile topological continuity. 
To solve these issues, we propose TopoTTA, which consists of two stages:  
\textit{Stage~1} performs topological structure adaptation to overcome the insufficient cross-domain topological structure adaptation; 
\textit{Stage~2} conducts topological continuity refinement to enhance the topological continuity of the final predictions. 
An overview of our TopoTTA is illustrated in~\figref{fig:main}. 
The details of \textit{Stage~1} and \textit{Stage~2} are described in Section~\ref{sec:stage1} and Section~\ref{sec:stage2}, respectively.

\subsection{Stage~1: \stageone}\label{sec:stage1}

This stage aims to adapt the pre-trained TSS model to discrepant topological structures caused by domain shifts. 
To achieve this, we propose TopoMDCs based on diverse fundamental topological patterns and adaptively reweight and combine TopoMDCs according to the topological features of different regions. We introduce the details of TopoMDCs and the model parameter updating process below.

\vspace{1mm}\noindent
\textbf{Topology-Meta Difference Convolutions (TopoMDCs).}
TopoMDCs draw inspiration from central difference convolution (CDC)~\cite{cdc}, which captures the difference between the central pixel of a convolution kernel and its neighbors to detect local gradient changes. However, the local topological shapes of tubular structures often exhibit directionality and continuity. CDC, initially designed for other fine-grained vision tasks (e.g., face anti-spoofing~\cite{cdc,xun2025tpami,mmdg2024}), lacks the ability to perceive these two features of tubular structures, making them ineffective in representing the complex topologies and geometric constraints within local regions.
Let $\boldsymbol{x}_{\text{in}}$ denote the input feature map. The vanilla convolution  $\mathcal{C}_{0}$ and CDC $\mathcal C_c$ can be formulated as:
\begin{equation}
    \footnotesize
    \begin{split}
    &\mathcal C_0(r_x, r_y) =\!\!\!\!\!\!\!\!\!\!\sum_{(\Delta r_x,\Delta r_y) \in \mathcal R} \!\!\!\!\!\!\!\!\!\!w(\Delta r_x, \Delta r_y) \cdot \boldsymbol x_{\text{in}}(r_x - \Delta r_x, r_y-\Delta r_y),\\
    &\mathcal C_c(r_x, r_y) =\!\!\!\!\!\!\!\!\!\!\sum_{(\Delta r_x, \Delta r_y) \in \mathcal R} \!\!\!\!\!\!\!\!\!\!w(\Delta r_x, \Delta r_y) \cdot \boldsymbol x_{\text{in}}(r_x, r_y),\\
    \end{split}
    \label{eq:3}
\end{equation}
where $\mathcal{R} = \{(1,1), (0,1), \cdots, (-1,0), (-1,-1)\}$ represents the local receptive field of a $3\!\times\!3$ convolution kernel, $w$ denotes the parameters of the vanilla convolution layer, and $(r_x, r_y)$ represents the relative position within the convolution operation centered on the current pixel of $\boldsymbol{x}_{\text{in}}$.

To overcome this limitation of CDC, the proposed TopoMDCs perform a two-pixel differential operation by extending their reach to the eight neighboring directions beyond the center pixel, as shown in ~\figref{fig:main}~(a).
This design simulates various basic topological patterns, effectively accommodating the elongated and diverse trajectories of tubular structures and enhancing the model's capability to recognize various tubular objects. 
Let $\boldsymbol{x}_\text{in} \in \mathbb R^{H_{\text{in}}\times W_{\text{in}} \times C_{\text{in}}}$ denote the input feature map, the pre-trained vanilla convolution be $\mathcal{C}_{0}$, and the set of TopoMDCs be $\{\mathcal{C}_{1}, \mathcal{C}_{2}, \cdots, \mathcal{C}_{8}\}$. 

Taking $\mathcal{C}_1$, which extends toward the top-left pixel (see Fig.~\ref{fig:main}~(a)), as an example, it is defined as:
\begin{equation}
    \footnotesize
    \begin{split}
    \mathcal C_1(r_x, r_y) &= \mathcal C_c(r_x, r_y) - \!\!\!\!\!\!\!\!\!\!\sum_{(\Delta r_x,\Delta r_y) \in \mathcal R_1} \!\!\!\!\!\!\!\!\!\!w(\Delta r_x, \Delta r_y) \cdot \boldsymbol x_{\text{in}}(r_x, r_y)  \\
     &+\!\!\!\!\!\!\!\!\!\!\sum_{(\Delta r_x,\Delta r_y) \in \mathcal R_1} \!\!\!\!\!\!\!\!\!\!w(\Delta r_x, \Delta r_y) \cdot \boldsymbol x_{\text{in}}(r_x + 1, r_y + 1),
    \end{split}
    \label{eq:4}
\end{equation}
where $\mathcal R_1 = \{(-1,-1),(-1,0),(0,-1)\}$ is the local receptive field of $\mathcal C_1$. Due to space limitations, the formal definitions of other TopoMDCs are provided in Appendix.

TopoMDCs do not introduce any additional parameters for any convolution layer but instead, adaptively transform the vanilla convolution in $\mathcal F(\cdot; \theta)$. Specifically, we replace all $3\!\times\!3$ convolution layers in the encoder of $\mathcal F(\cdot; \theta)$ with TopoMDCs. 
All TopoMDCs inherit the parameters of the vanilla convolutions.
To adapt to a variety of complex (even unseen) topologies, a dynamic combination of TopoMDCs is required. 
Given that the topologies of local regions tend to be quite similar, while the topologies of distant regions can differ significantly, we introduce different learnable parameter sets $\boldsymbol\delta$ with eight learnable router parameters for different image patches.
To adapt to the respective topological structures in different regions, we patchify the input image into $n \times n$ non-overlap patches, each with the size of $H_{\text{in}}/n \times W_{\text{in}}/n$.
Taking the $j$-th image patches $\boldsymbol p_j$ from $\boldsymbol{x}_{\text{in}}$, the corresponding router parameter set $\boldsymbol\delta$ assigns external parameters to different TopoMDCs, enabling adaptive enhancement of the perception of topological structures within the patch.
The output feature $\boldsymbol{\hat{x}}_{\text{out}}$ is  obtained as: 
\begin{equation}
    \small
    \boldsymbol{\hat{x}}_{\text{out}} =\mathcal{C}_0(\boldsymbol{x}_{\text{in}}) - \sum_{j=1}^{n \times n} \boldsymbol \delta_j \sum_{i=1}^8  \mathcal{C}_i(\boldsymbol{x}_{\text{in}}^j),
    \label{eq:5}
\end{equation}
where $\boldsymbol{x}_{\text{in}}^j$ denotes the $j$-th patch of $\boldsymbol{x}_{\text{in}}$. The patchification operations $\mathcal{C}_0$ and $\mathcal{C}_i$ differ slightly in their inputs in ~\eqnref{eq:4} and ~\eqnref{eq:5}. In ~\eqnref{eq:5}, $\mathcal{C}_0$ and $\mathcal{C}_i$ denote the same computation applied uniformly to every pixel in $\boldsymbol{x}_{\text{in}}^j$.

\vspace{1mm}\noindent
\textbf{Router Parameter ($\boldsymbol \delta$) Updating in Stage 1.}
As mentioned above, we do not update the original parameters of the convolution kernel or any other parameters in the network but only update a small number of introduced router parameters $\boldsymbol \delta$. This is not just a consideration for inference speed. 
Simultaneously adjusting both $\boldsymbol \delta$ and the model parameters increases the search-space complexity, which may lead to conflicts or interference among parameters. 
This, in turn, could even disrupt the original feature representation.
To this end, we only adjust $\boldsymbol \delta$ to adapt the topological response within an external feature space, thereby reducing the risk of instability.
For each incoming test sample $\boldsymbol{x}$, all router parameters are reset to zero, so that the model can independently adapt to the topological structures of the new sample.
We use the self-supervised entropy minimization (EM) loss~\cite{tta_tent} to update $\boldsymbol \delta$:
\begin{equation}
    \small
    \begin{split}
    &\mathcal{L}_{\text{EM}} = -\sum_{j=1}^2 \boldsymbol{\hat{y}}_j \log(\boldsymbol{\hat{y}}_j),\\
    &\boldsymbol \delta = \mathop{\arg\min}\limits_{\theta^t}\mathcal L_{\text{EM}}(\mathcal F(\cdot; \theta^t;\boldsymbol \delta), \boldsymbol x).
    \end{split}
\end{equation}

\subsection{Stage 2: \stagetwo}\label{sec:stage2}

As previously discussed in Sec.~\ref{sec:stage1}, \textit{Stage~1} enhances the representation of topological structures in $\mathcal{D}^{t}$.
In \textit{Stage~2}, we guide the model to learn local characteristics in $\mathcal{D}^{t}$ for foreground-background differentiation, thereby improving the topological continuity of segmentation.
To achieve this, we generate a series of \textit{pseudo-breaks} at key topological structures on the test image $\boldsymbol x$, producing the edited image $\boldsymbol{x}^{\prime}$ through the proposed Topology Hard sample Generation (TopoHG) strategy, as shown in \figref{fig:main}~(c). 

\vspace{1mm}\noindent
\textbf{Topology Hard Sample Generation (TopoHG).}
The TopoHG process for generating local hard samples can be roughly divided into the following three steps:

\vspace{0.5mm}
\textit{Step~1:~Key Point Selection.} Since we apply consistency supervision on the predictions of local hard samples, the selected regions must have highly reliable pseudo-labels. 
To this end, we first select a set of points with confidence above the threshold $\tau = 0.95$ from the teacher's predictions $ \boldsymbol{\hat{y}}^{\prime}$, denoted as $\mathcal{P} = \{(u, v) \mid \boldsymbol{\hat{y}}^{\prime}(u, v) > \tau\}$. 
Then, we randomly select $N_p$ points from $\mathcal{P}$ as key points. Since prediction confidence varies across different test samples, using a fixed $N_p$ for all input samples is suboptimal. 
Therefore, we allow samples with higher confidence to select more key points and create more challenging regions to contribute more to the weight update: $N_p = k \cdot |\mathcal{P}|$, where $k$ is a hyperparameter.

\vspace{0.5mm}
\textit{Step~2:~Sliding to Search. }
For each selected key point $(u_c, v_c)$, let $\boldsymbol x_p^{\text{fg}}$ denote the fixed foreground window centered at this point, with a size of $s\!\times\!s$. Then, a background sliding window $\boldsymbol x_p^{\text{bg}}$, of the same size as the foreground window, slides over adjacent non-overlapping regions, identifying window $\boldsymbol x_p^{\text{bg},*}$ with the lowest pseudo-label confidence of all pixels. Meanwhile, to prevent $\boldsymbol x_p^{\text{bg},*}$ from carrying too much foreground information, we set a lower threshold $\tau^{\text{bg}}$. If the sum of foreground confidence of all pixels in $\boldsymbol x_p^{\text{bg},*}$ exceeds $\tau^{\text{bg}}$, the key point $(u_c, v_c)$ is discarded.

\vspace{0.5mm}
\textit{Step~3:~Frequency-based \textit{Pseudo-break} Generation.}
TopoHG aims to create local \textit{pseudo-break} topological structures within the critical regions. 
This requires TopoHG to simulate the characteristics of structures that seem to be broken while retaining essential foreground features. 
Inspired by~\cite{douqi_freq}, we adopt a low-frequency swap method between background and foreground to preserve high-frequency information that carries critical foreground features. 
Let $\text{FFT}$ and $\text{iFFT}$ denote the Fast Fourier Transform (FFT) and inverse Fast Fourier Transform (iFFT) operations~\cite{fft}, respectively, with $\boldsymbol m_{\text{low}}$ as the mask for the low-frequency regions that need to be swapped. Applying FFT to $\boldsymbol{x}_{p}^{\text{fg}}$ and $\boldsymbol{x}_{p}^{\text{bg,*}}$ yields the frequency domain representations $\boldsymbol{f}_{p}^{\text{fg}} = \text{FFT}(\boldsymbol{x}_{p}^{\text{fg}})$ and $\boldsymbol{f}_{p}^{\text{bg}} = \text{FFT}(\boldsymbol{x}_{p}^{\text{bg,*}})$. After performing low-frequency swap on $\boldsymbol{f}_{p}^{\text{fg}}$ and $\boldsymbol{f}_{p}^{\text{bg}}$ and applying iFFT, we obtain the modified image $\boldsymbol{x}_{p}^{\text{swap}}$, as follows:
\begin{equation}
    \small
    \boldsymbol{x}_{p}^{\text{swap}} = \text{iFFT}\left(\boldsymbol{f}_{p}^{\text{fg}} \cdot \left(1-\boldsymbol m_{\text{low}}\right) + \boldsymbol{f}_{p}^{\text{bg}} \cdot \boldsymbol m_{\text{low}}\right).
    \label{eq:7}
\end{equation}
To avoid additional risks associated with unnecessary background changes when emphasizing consistency, we update only update pixels in $\boldsymbol{x}_{p}^{\text{fg}}$ that correspond to the foreground in the pseudo-label $\boldsymbol{\hat{y}}_{p}^{\prime}$. The \textit{pseudo-break} patch $\boldsymbol{x}_{p}^{\prime}$ can be defined as follows:
\begin{equation}
    \small
    \boldsymbol{x}_{p}^{\prime} = \boldsymbol{x}_{p}^{\text{swap}} \cdot \boldsymbol{\hat{y}}_{p}^{\prime} + \boldsymbol{x}_{p}^{\text{fg}} \cdot (1 - \boldsymbol{\hat{y}}_{p}^{\prime}).
    \label{eq:8}
\end{equation}
After completing the above four steps, we obtain a hard sample $\boldsymbol{x}^{\prime}$ that contains multiple local \textit{pseudo-breaks}, resulting in weakened predictions to assist in model updates.

\begin{table*}[t]
	\caption{Cross-domain testing results in four different scenarios, i.e., retinal vessel segmentation, road extraction, microscopic neuronal segmentation, and retinal OCT-angiography vessel segmentation. Source Only: Trained on the source domain and tested on the target domain directly. The best and second-best results in each column are highlighted in \textbf{bold} and \underline{underline}, respectively.}
        \vspace{-10pt}
	\setlength{\tabcolsep}{5pt}
        \renewcommand\arraystretch{0.78}
	\resizebox*{1.0 \linewidth}{!}{
		\begin{tabular}{lccccccccccccccc}
			\toprule[1.2pt]
			\multirow{2}{*}{\textbf{Method}} & \multicolumn{3}{c}{\textbf{DRIVE $\rightarrow$ CHASE}} &  \multicolumn{3}{c}{\textbf{DRIVE $\rightarrow$ STARE}} &  \multicolumn{3}{c}{\textbf{CHASE $\rightarrow$ DRIVE}} &  \multicolumn{3}{c}{\textbf{CHASE $\rightarrow$ STARE}} &  \multicolumn{3}{c}{\textbf{DeepGlobe $\rightarrow$ MR}} \\
			\cmidrule(lr){2-4}\cmidrule(lr){5-7}\cmidrule(lr){8-10}\cmidrule(lr){11-13}\cmidrule(lr){14-16}
			& Dice (\%) $\uparrow$ & clDice (\%) $\uparrow$ & $\beta$ $\downarrow$ & Dice (\%) $\uparrow$ & clDice (\%) $\uparrow$ & $\beta$ $\downarrow$ & Dice (\%) $\uparrow$ & clDice (\%) $\uparrow$ & $\beta$ $\downarrow$ & Dice $\uparrow$ & clDice (\%) $\uparrow$ & $\beta$ $\downarrow$ & Dice (\%) $\uparrow$ & clDice (\%) $\uparrow$ & $\beta$ $\downarrow$ \\
                \midrule[0.7pt]

			Source Only & 22.38  & 18.21  &  41.25  & 48.05  &  /  & 106.30  &  62.95  & 58.59  &  89.95 &  48.77  &  /  &  105.20  &  44.89 &  54.97 & 76.83 \\
			TENT \cite{tta_tent} & 66.97  & 69.43  &  37.75  & 67.14  &  60.74  & 98.40  &  64.74  & 60.25  &  88.30 &  61.91  &  55.39  &  106.00  &  42.61 &  52.18 & 77.71 \\
                CoTTA \cite{tta_cotta} & \underline{68.60}  & \underline{71.53}  &  \underline{36.38}  & 67.02  &  59.86  & 97.40  &  \underline{67.64}  & \underline{64.80}  &  \underline{81.20} &  \underline{63.72}  &  \underline{57.64}  &  \underline{99.85}  &  \textbf{50.83} & \underline{60.58}  & \underline{74.82}\\
			SAR \cite{tta_SAR} & 66.97  & 69.49  &  37.13  & \underline{67.29}  & 60.90 & 97.90  &  65.06  & 60.69  &  88.35 & 62.17  &  55.68  &  105.60  &  43.88 & 54.30  & 76.70 \\
                DIGA \cite{tta_diga} & 66.91  & 70.37  & 36.50  & 59.22  &  \underline{61.77}  & \textbf{82.00}  & 66.54  & 62.43  & 85.85 & 63.80 & 57.32  & 103.75 & 40.65  &  51.08 & 78.82 \\
			DomainAdaptor \cite{tta_domainadaptor} & 64.99  & 66.99  & 41.50  & 62.98  &  55.51  & 102.75  & 64.43  & 59.89  & 88.50 & 59.41  &  52.05  &  105.45  & 45.17  &  55.86 & 76.98 \\
			MedBN \cite{tta_medbn} & 49.77  & 51.09  &  75.88  & 57.83  &  51.99  & 102.55  &  59.63  & 54.14  &  89.95 & 54.05  &  48.27  &  109.00  & 38.92  &  / & 76.89 \\
			VPTTA \cite{tta_vptta} & 66.48  & 68.93  &  38.50  & 67.16  &  60.54  & 98.95  &  65.02  & 60.70  &  88.35 & 61.73  &  55.01  &  105.55  &  44.51 &  55.99 & 77.43  \\
                \midrule
                \rowcolor{gray!20}
			\textbf{\modelName~(Ours)} & \textbf{70.73}  & \textbf{77.05}  &  \textbf{25.38} &  \textbf{67.36}  & \textbf{62.74}  &  \underline{85.75}  & \textbf{72.96}  &  \textbf{70.26}  &  \textbf{79.15}  &  \textbf{68.43}  &  \textbf{61.20}  &  \textbf{97.60}  &  \underline{50.74} &  \textbf{66.08} & \textbf{69.23} \\
                \midrule[1.0pt]
                \multirow{2}{*}{\textbf{Method}} & \multicolumn{3}{c}{\textbf{DeepGlobe $\rightarrow$ CNDS}} &  \multicolumn{3}{c}{\textbf{Neub1 $\rightarrow$ Neub2}} &  \multicolumn{3}{c}{\textbf{Neub2 $\rightarrow$ Neub1}} &  \multicolumn{3}{c}{\textbf{ROSE $\rightarrow$ OCTA500}} &  \multicolumn{3}{c}{\textbf{OCTA500 $\rightarrow$ ROSE}} \\
			\cmidrule(lr){2-4}\cmidrule(lr){5-7}\cmidrule(lr){8-10}\cmidrule(lr){11-13}\cmidrule(lr){14-16}
			& Dice (\%) $\uparrow$ & clDice (\%) $\uparrow$ & $\beta$ $\downarrow$ & Dice (\%) $\uparrow$ & clDice (\%) $\uparrow$ & $\beta$ $\downarrow$ & Dice (\%) $\uparrow$ & clDice (\%) $\uparrow$ & $\beta$ $\downarrow$ & Dice (\%) $\uparrow$ & clDice (\%) $\uparrow$ & $\beta$ $\downarrow$ & Dice (\%) $\uparrow$ & clDice (\%) $\uparrow$ & $\beta$ $\downarrow$ \\
                \midrule[0.7pt]
            
			Source Only  &  \underline{81.74} & \underline{92.22}  & 7.09 & 14.34  & /  &  8.22  & 60.97  &  65.62  & 60.56  &  50.10  & 57.65  &  55.36 &  71.39  & 74.64  &  19.55\\
			TENT \cite{tta_tent} &  78.69 &  90.65 & 6.65 & 51.55  & 60.58  &  9.06  & 63.26  &  72.32  & 9.37  &  68.99  & \underline{75.69}  &  43.06 &  72.25 &  75.64 &  17.89\\
                CoTTA \cite{tta_cotta} &  77.83 &  91.18 & \underline{5.99} & 52.21  & 62.78  &  10.06  & \textbf{64.40}  &  \underline{73.58}  & \underline{7.96}  &  \underline{70.05}  & 75.65  &  \underline{41.34} &  72.58 & 75.78  & 18.00 \\
			SAR \cite{tta_SAR} & 79.19  & 90.72  & 6.86 & 51.85  & 60.80  & 9.16  & 63.21  & 72.31 & 9.62  &  67.65  & 74.14  &  44.42 & 72.45 &  75.85 &  17.44\\
                DIGA \cite{tta_diga} & 75.45  & 83.22  & 14.10  & \underline{59.14}  &  \underline{66.44}  & 19.77  & 61.32  & 73.40  & 8.19 & 66.17  & 72.98  & 42.18 & 71.51  &  75.40 & 17.99 \\
			DomainAdaptor \cite{tta_domainadaptor} &  80.71 &  91.72 & 6.80 & 53.24  & 60.96  & 7.94  & \underline{64.04}  &  72.38  & 8.75  & 67.39  & 74.60  & 44.14 & 71.77 &  75.37 & 17.00\\
			MedBN \cite{tta_medbn} & 76.96  & 87.94  & 9.22 & 55.59  & 63.92  &  10.56  & 61.34  &  70.82  & 11.57  &  40.64  & 47.53  &  298.74 & 71.77 & 73.29  & \textbf{16.11} \\
			VPTTA \cite{tta_vptta} & 77.01  &  88.56 & 8.64 & 54.25  & 62.88  &  \underline{8.83}  & 63.46  &  72.24  & 9.34  &  67.54  & 74.01  &  44.92 & \underline{72.61} &  \underline{75.93}  & 17.11 \\
                \midrule
                \rowcolor{gray!20}
			\textbf{\modelName~(Ours)} & \textbf{89.15}  &  \textbf{96.35} & \textbf{5.98} & \textbf{66.88}  & \textbf{74.95}  &  \textbf{7.22}  &  61.83  &  \textbf{75.38}  & \textbf{6.50}  &  \textbf{70.70}  & \textbf{78.24}  &  \textbf{31.88}  & \textbf{75.59} &  \textbf{77.73} &  \underline{16.22} \\
			\bottomrule[1.2pt]
		\end{tabular}
	}
    \vspace*{-12pt}
    \label{tab:p1}
\end{table*}

\vspace{1mm}\noindent
\textbf{Model Updating in Stage 2.}
To align the predictions $\boldsymbol{\hat{y}}^{\prime\prime}$ of these local hard samples $\boldsymbol{x}^{\prime}$ with the original pseudo-label $\boldsymbol{\hat{y}}^{\prime}$, we apply consistency regularization as supervision, implemented through the TS scheme as shown in \figref{fig:main}~(c). The cross-entropy loss $\mathcal{L}_{\text{CE}}$ is used to enhance the consistency between the student's (i.e., prediction of the hard sample $\boldsymbol{\hat{y}}^{\prime\prime}$) and teacher's (i.e., pseudo-label $\boldsymbol{\hat{y}}^{\prime}$ of original samples) predictions, updating $\theta_i^t$ as follows:
\begin{equation}
    \small
    \begin{split}
        \mathcal{L}_{\text{CE}} &= -\sum_{(u, v)} \mathcal{W}\left(u, v\right) \cdot \left(\boldsymbol{\hat{y}}^{\prime} \log\left(\boldsymbol{\hat{y}}^{\prime\prime}\right) +  \boldsymbol{\hat{y}}^{\prime\prime} \log\left(\boldsymbol{\hat{y}}^{\prime}\right)\right),\\
    \theta^{t}_i &= \mathop{\arg\min}\limits_{\theta^t_{i-1}} \mathcal {L}_{\text{CE}}(\mathcal F(\cdot; \theta^t_{i-1}; \mathbf{\delta}), \boldsymbol x ),
    \end{split}
\end{equation}
where $\mathcal{W}(u, v)$ is a weight map. The local \textit{pseudo-breaks} should receive higher attention in consistency regularization; therefore, $\mathcal{W}(u, v)$ is defined as follows:
\begin{equation}
    \small
    \mathcal{W}(u, v) = 
    \begin{cases} 
        10 & \text{if } (u, v) \in \boldsymbol{x}_{p}^{\prime} \cdot \boldsymbol{\hat{y}}_{p}^{\prime} \\ 
        1 & \text{otherwise}
    \end{cases}.
\end{equation}
The gradient is used solely to update the student, while the teacher's parameters are updated by an exponential moving average (EMA) of the student's parameters. Finally, after several steps of updates, the unmodified input $\boldsymbol{x}$ is fed into the student model to obtain the final prediction.

\section{Experiments}

We conduct comparison experiments across four scenarios, i.e., retinal vessel segmentation, road extraction, microscopic neuronal segmentation, and retinal OCT-angiography vessel segmentation covering ten datasets. 

\subsection{Experiment Setup}
\label{sec:exp_setup}
\vspace{1mm}\noindent
\looseness=-1
\textbf{Datasets.}
We adopt 10 widely-used TSS datasets for the following experiments. The retinal vessel segmentation datasets are collected from three different sources: DRIVE~\cite{drive}, STARE~\cite{stare}, and CHASEDB1 (CHASE)~\cite{chase}, which contain 40, 20, and 28 samples, respectively.
Three road extraction datasets are DeepGlobe~\cite{deepglobe}, Massachusetts road (MR)~\cite{mr}, and CNDS~\cite{cnds}, providing 8,570, 1,171, and 224 samples, respectively. 
The microscopic neuronal segmentation datasets are Neub1 and Neub2~\cite{neub}, with 112 and 98 samples, respectively. 
Two retinal OCT-angiography vessel segmentation datasets are from different sources, ROSE1~\cite{rose} and OCTA500~\cite{octa}, containing 117 and 300 samples, respectively. 
All datasets follow their original training and testing splits, except for STARE, which is officially used as a test set due to its limited sample size~\cite{stare}.

\vspace{0.5mm}\noindent
\textbf{Evaluation Metrics.} Following prior works~\cite{qi2023dscnet, deepclose, pointscatter, dconnnet}, we use Dice as the primary metric for TSS performance evaluation, with clDice~\cite{cldice} and \textit{Betti errors} ($\beta$)~\cite{hu2019topology} as metrics of topological continuity. 
In \tabref{tab:p1}, the diagonal marks ``/'' in clDice for some TTA methods indicate cases where no skeleton map can be extracted from their predictions, and thus clDice cannot be calculated. 
Further investigation reveals that these are failure cases.

\vspace{0.5mm}\noindent
\textbf{Implementation Details.}
For each scenario, we use a model pre-trained on a single dataset (source domain) and test it on other datasets (target domains) to calculate metrics. 
UNet~\cite{unet} and CS2Net~\cite{cs2net} (designed for TSS) are adopted as baseline networks for validating the performance of TopoTTA. 
To ensure consistent field-of-view and cross-domain accuracy, all datasets are resized to $384\!\times\!384$ pixels, except for the OCTA dataset as ROSE and OCTA500 already share similar resolutions. 
During testing, TopoTTA and the comparison methods undergo six iterations of adaptation for each image, with TopoTTA performing three iterations at each stage to ensure experimental consistency. For TopoTTA, we use the Adam optimizer with learning rates of $0.01$ and $1\!\times\!10^{-4}$ for the first and second stages, respectively. We use 0.5 as the threshold for mask binarization. The hyperparameters $n\!\times\!n$ (number of TopoMDC regions), $s$ (modification window size), and $\tau^{\text{bg}}$ (upper limit of the foreground pixel), respectively, are set to $4\!\times\!4$, 30, 0.05 for all tasks. 
We present more details in Appendix.

\vspace{1mm}\noindent
\textbf{Comparison TTA Methods.}
In the following experiments, we compare our method with three categories of TTA methods: 
(1) methods designed for semantic segmentation, such as CoTTA~\cite{tta_cotta} and DIGA~\cite{tta_diga}, 
(2) a method designed for medical image segmentation, i.e., VPTTA~\cite{tta_vptta}, and
(3) methods designed for classification, including TENT~\cite{tta_tent}, SAR~\cite{tta_SAR}, DomainAdaptor~\cite{tta_domainadaptor} and MedBN~\cite{tta_medbn}, which can be easily adapted to segmentation tasks~\cite{tta_vptta, tta_diga} by using pixel-level self-supervision or BN statistics adjustment.
We implement these TTA methods on the same baseline networks as TopoTTA, using hyperparameters as specified in the corresponding reference or assigning optimal ones.

\begin{table}[t]
\setlength{\tabcolsep}{5pt}
\caption{Average cross-dataset testing results across ten datasets using UNet and CS2Net as baseline networks. The best and second-best results in each column are highlighted in \textbf{bold} and \underline{underline}, respectively.}
\vspace{-8pt}
\setlength{\tabcolsep}{7.9pt}
\renewcommand\arraystretch{0.82}
\resizebox*{1.0 \linewidth}{!}{
    \begin{tabular}{lcccccc}
    \toprule[1.2pt]
    \multirow{2}{*}{\textbf{Method}} & \multicolumn{3}{c}{\textbf{UNet}} & \multicolumn{3}{c}{\textbf{CS2Net}}\\
    \cmidrule(lr){2-4}\cmidrule(lr){5-7}
     & Dice (\%) $\uparrow$ & clDice (\%) $\uparrow$ & $\beta$ $\downarrow$ & Dice (\%) $\uparrow$ & clDice (\%) $\uparrow$ & $\beta$ $\downarrow$ \\
    \midrule[0.7pt]
    Source Only & 50.56 & 42.19 & 51.94  & 50.31 & 51.17  & 54.35     \\
    TENT~\cite{tta_tent} & 63.81  & 67.29  & 49.42 & 63.76  & 68.09  &  51.87    \\
    CoTTA~\cite{tta_cotta} & \underline{65.49}  & \underline{69.34} & \underline{47.30}  &  \underline{65.12}  & \underline{69.84}  & \underline{48.19}    \\
    SAR~\cite{tta_SAR} & 63.97  & 67.49 & 49.32 & 63.63  & 68.11  &  51.84  \\
    DIGA~\cite{tta_diga} & 63.07  & 67.44 & 48.92 & 63.92  & 68.20  &  51.97  \\
    DomainAdaptor~\cite{tta_domainadaptor} & 63.41 & 66.53 & 49.98 & 60.95  & 65.65  &  54.68    \\
    MedBN~\cite{tta_medbn} & 56.65 & 54.90 & 80.05 & 62.63  & 65.97  & 54.63   \\
    VPTTA~\cite{tta_vptta} & 63.98 & 67.48 & 49.76 & 63.30  & 67.77  &  52.25  \\
    \midrule
    \rowcolor{gray!20}
    \textbf{\modelName~(Ours)} & \textbf{69.44} & \textbf{74.00} & \textbf{43.01} & \textbf{67.68} & \textbf{72.70} &  \textbf{43.86} \\
    \bottomrule[1.2pt]
    \end{tabular}
}
\label{tab:p2}
\vspace{-15pt}
\end{table}

\begin{figure*}[t]
	\setlength{\abovecaptionskip}{1pt}
	\setlength{\belowcaptionskip}{-16pt}
	\centering 
	\includegraphics[width=\textwidth]{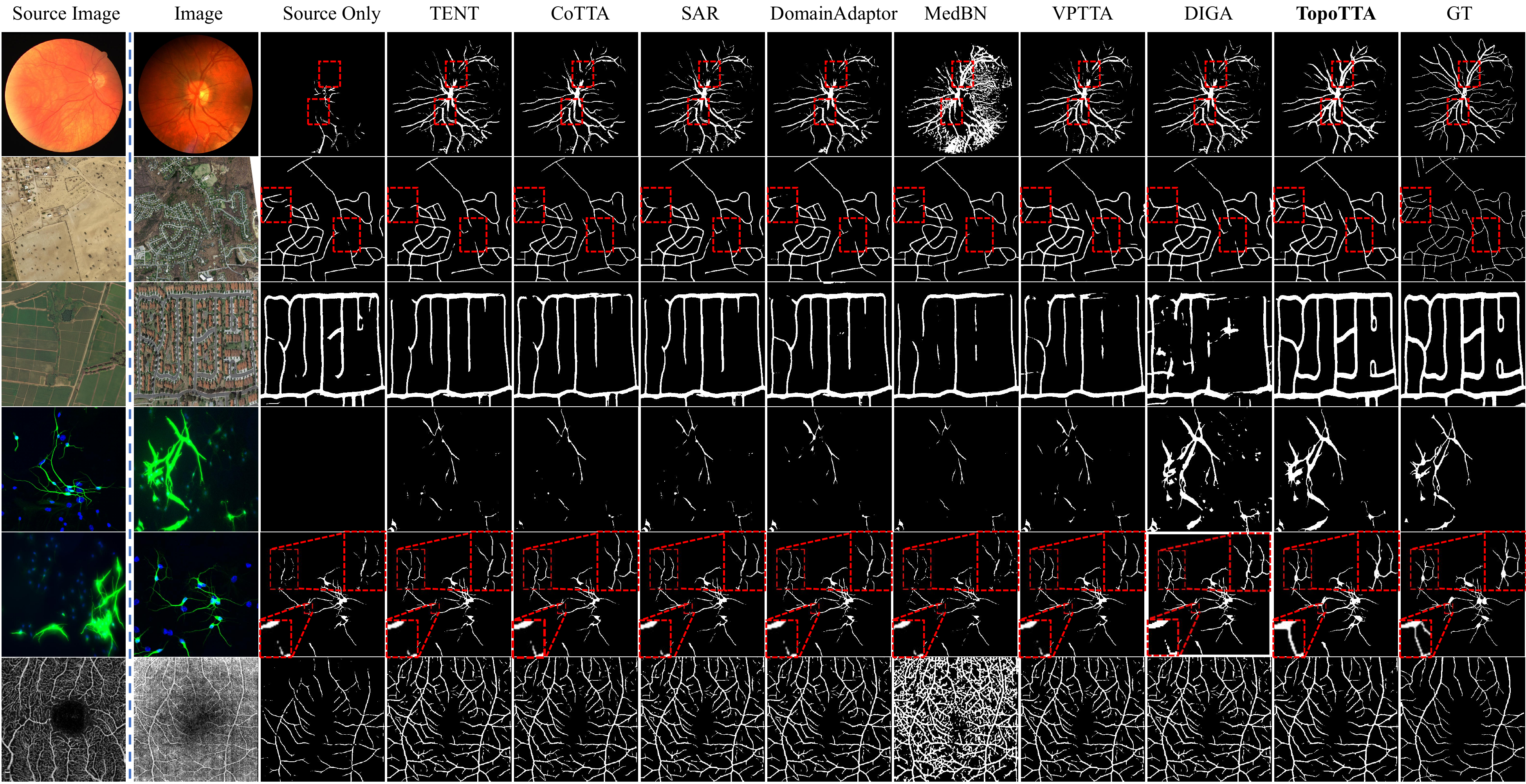}
	\caption{Visualization of segmentation results for TopoTTA and seven comparison methods across four scenarios. ``Source Image'' is provided solely to illustrate domain differences across datasets. ``Source Only'' denotes results without any TTA methods applied, and GT is short for ground-truth labels. More visual examples are presented in Appendix.} 
	\label{fig:visual}
\end{figure*}

\subsection{Comparison Results}

\tabref{tab:p1} presents the cross-dataset testing results across four different scenarios. \tabref{tab:p2} shows the average results for each baseline across all scenarios. \figref{fig:visual} visualizes the segmentation results. From these results, we have the following key observations:

\vspace{1mm}\noindent
\textbf{Key Observation 1: Improvement on topological continuity across four scenarios.}
As shown in \tabref{tab:p1}, TopoTTA consistently achieves the highest topological continuity metrics. This improvement is not attained at the expense of segmentation accuracy (e.g., DIGA in the DRIVE $\rightarrow$ STARE, and VPTTA in the DeepGlobe $\rightarrow$ MR), but rather, TopoTTA enhances Dice score on most datasets. Notably, for the DeepGlobe $\rightarrow$ CNDS experiment, where the source domain model has already demonstrated strong test time performance, the results of all comparison methods decline. However, TopoTTA substantially improves segmentation accuracy and topological continuity metrics in this experiment. In particular, in the challenging Neub1 $\rightarrow$ Neub2 experiment, TopoTTA surpasses the second-best method, achieving gains of 7.74 and 8.51 in Dice and clDice scores, respectively, underscoring the method’s exceptional capacity for topological adaptation and continuity enhancement. Additionally, in the Neub1 $\rightarrow$ Neub2 experiment, although TopoTTA’s Dice score is lower than most comparison methods, visual analysis indicates this is likely due to label granularity rather than a true performance drop. As space is limited, details are provided in Appendix.

\vspace{0.5mm}\noindent
\textbf{Key Observation 2: Consistent superiority across different baselines.}
TopoTTA can function as a plug-and-play module within standard convolutional neural networks. From \tabref{tab:p2}, we observe that TopoTTA achieves significant improvements across ten experiments using both UNet and CS2Net as baseline networks. CoTTA attains the second-best performance, indicating that the TS scheme may be well-suited for segmentation tasks. With UNet as the baseline network, TopoTTA outperforms CoTTA by 3.95\% in Dice, 4.66\% in clDice, and reduces \textit{Betti errors} by 4.29\%. Similar gains are observed using CS2Net. We also implement a TopoTTA variant on DSCNet, which delivers notable performance improvements, demonstrating TopoTTA’s adaptability across different baseline networks. Detailed results on other baseline networks (i.e., CS2Net and DSCNet) can be found in Appendix.

\vspace{0.5mm}\noindent
\textbf{Key Observation 3: Clear visual proof of enhanced continuity and topological adaptability.}
In \figref{fig:visual}, we magnify specific regions of selected images for detailed comparison. In rows 1, 2, and 5, all comparison methods show apparent discontinuities in the red-boxed areas, while TopoTTA effectively connects these branches, underscoring its superior continuity enhancement. For the road image in row 3, TopoTTA achieves an almost perfect match with the ground truth---a level of accuracy beyond all comparison methods' reach. In the microscopic neuronal image in row 4, most comparison methods struggle to segment even the main branches, while TopoTTA closely matches the ground truth. These results showcase TopoTTA's adaptability to various topological features.
Moreover, in row 6, comparison methods exhibit noise due to over-segmentation, which TopoTTA effectively suppresses.

\begin{table}[t]
\setlength{\tabcolsep}{5pt}
\caption{Ablation results of the proposed TopoTTA. Baseline\textsuperscript{$\star$} is the model using the teacher-student scheme.}
\vspace{-10pt}
\setlength{\tabcolsep}{7.9pt}
\renewcommand\arraystretch{0.82}
\resizebox*{1.0 \linewidth}{!}{
    \begin{tabular}{lccc}
    \toprule[1.2pt]
    {\textbf{Method}} & Dice (\%) $\uparrow$ & clDice (\%) $\uparrow$ & $\beta$ $\downarrow$ \\
    \midrule[0.7pt]
    Baseline & 65.37 & 61.69  & 82.24 \\
    Baseline + TopoMDCs (\textit{Stage 1}) & 68.70  & 65.14  & 76.28 \\
    Baseline\textsuperscript{$\star$} & 65.95 & 62.20 & 77.90 \\
    Baseline\textsuperscript{$\star$} + TopoHG (\textit{Stage 2}) & 68.82  & 66.61 & 73.63 \\
    \midrule
    \rowcolor{gray!20}
    \textbf{\modelName} & \textbf{69.87} & \textbf{67.81} & \textbf{73.27} \\
    \bottomrule[1.2pt]
    \end{tabular}
}
\label{tab:abla_main}
\vspace{-10pt}
\end{table}

\subsection{Ablation Study}

In this section, we perform ablation analyses of TopoMDCs and TopoHG. 
For all ablation experiments, we report the average results on the retinal vessel segmentation scenario.
Due to the limited space, the ablation analysis on our hyperparameters is provided in Appendix.

\begin{figure}[t]
    \setlength{\abovecaptionskip}{1pt}
	\setlength{\belowcaptionskip}{-10pt}
	\centering 
	\includegraphics[width=0.49\textwidth]{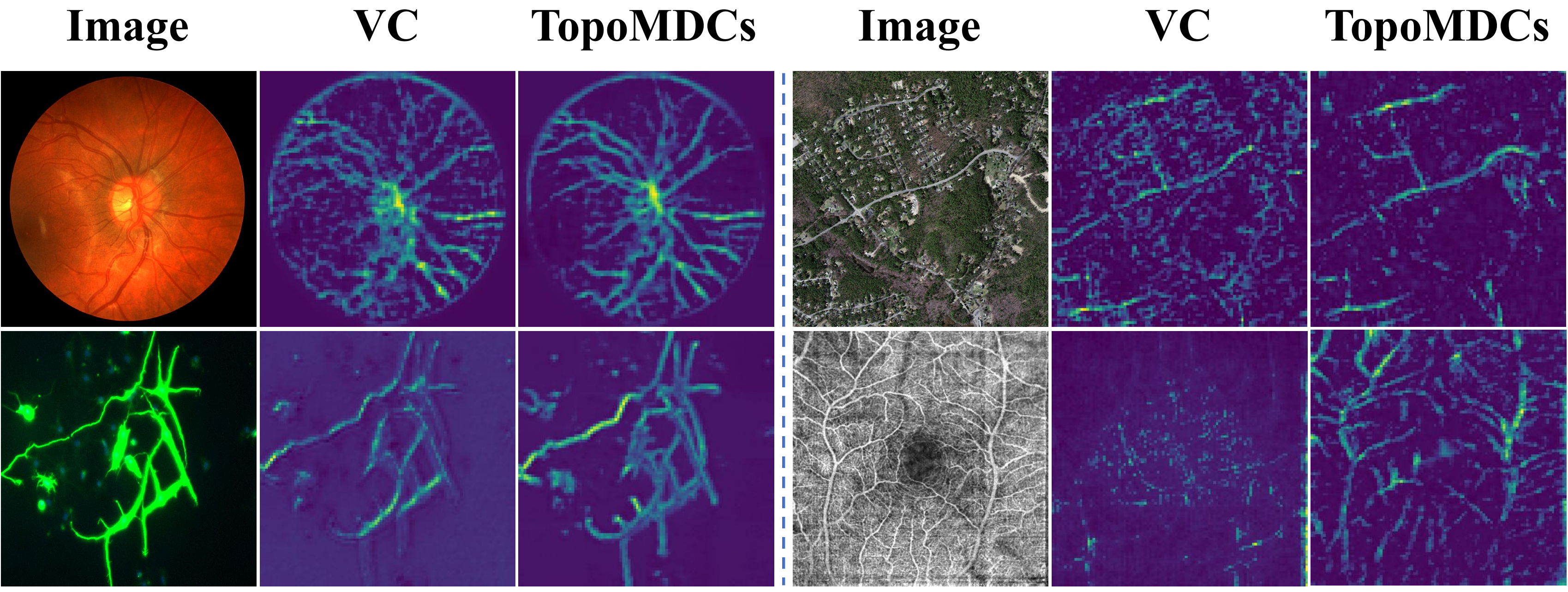}
	\caption{Feature maps before and after TopoMDCs across four scenarios. VC represents the feature maps of vanilla convolutions.} 
	\label{fig:feature_map} 
\end{figure}

\vspace{1mm}\noindent
\looseness=-1
\textbf{Effectiveness of TopoMDCs.}
To validate the effectiveness of the proposed TopoMDCs, we compare the network performance before and after applying TopoMDCs to the baseline or removing them from the TopoTTA. 
As shown in \tabref{tab:abla_main}, TopoMDCs result in significant improvements across all metrics compared to the baseline that only updates statistical parameters, validating the effectiveness of TopoMDCs in adapting to various topological structures. Additionally, \tabref{tab:abla_mdc} provides a detailed analysis of the contribution of each type of TopoMDCs to performance enhancement. We categorize TopoMDCs into three types: central ($\mathcal{C}_c$), orthogonal ($\mathcal{C}_{1-4}$), and diagonal ($\mathcal{C}_{5-8}$), and evaluate their impact individually. Results indicate that orthogonal and diagonal types relative to central lead to more significant performance improvements, with optimal topological awareness achieved when orthogonal and diagonal types are used together. We also visualize feature maps before and after incorporating TopoMDCs. Fig.~\ref{fig:feature_map} illustrates that in all four scenarios, TopoMDCs effectively enhance the topological structures on the target domain.

\begin{table}[t]
\setlength{\tabcolsep}{5pt}
\caption{Ablation results of the proposed TopoMDCs using each TopoMDC type individually.}
\vspace{-10pt}
\setlength{\tabcolsep}{7.9pt}
\renewcommand\arraystretch{0.82}
\resizebox*{1.0 \linewidth}{!}{
    \begin{tabular}{@{}llccc@{}}
    \toprule
    \textbf{Variant}         & \textbf{MDC Operation Set}                                                              & Dice (\%) $\uparrow$ & clDice (\%) $\uparrow$ & $\beta$ $\downarrow$ \\ \midrule
    CDC             & $\{\mathcal{C}_{c}$\}                                                    & 68.99                & 67.35                  & \textbf{72.91}       \\
    Orthogonal MDCs          & $\{\mathcal{C}_{1}, \mathcal{C}_{2}, \mathcal{C}_{3}, \mathcal{C}_{4}$\} & 69.26                & 67.54                  & 72.93                \\
    Diagonal MDCs            & $\{\mathcal{C}_{5}, \mathcal{C}_{6}, \mathcal{C}_{7}, \mathcal{C}_{8}$\} & 69.29                & 67.48                  & 73.86                \\
    \midrule
    \rowcolor{gray!20}
    \textbf{TopoMDCs (Full)} & $\{\mathcal{C}_{1}, \mathcal{C}_{2},\cdots, \mathcal{C}_{8}$\}           & \textbf{69.87}       & \textbf{67.81}         & 73.27                \\ \bottomrule
    \end{tabular}
}
\label{tab:abla_mdc}
\vspace{-10pt}
\end{table}

\begin{table}[t]
\setlength{\tabcolsep}{5pt}
\caption{Ablation results of the proposed TopoHG. We compare with three variants, i.e., Gaussian blur, random Gaussian noise, and image swap in the spatial domain.}
\vspace{-10pt}
\setlength{\tabcolsep}{20pt}
\renewcommand\arraystretch{0.82}
\resizebox*{1.0 \linewidth}{!}{
    \begin{tabular}{lccc}
    \toprule[1.2pt]
    {\textbf{Method}} & Dice (\%) $\uparrow$ & clDice (\%) $\uparrow$ & $\beta$ $\downarrow$ \\
    \midrule[0.7pt]
    Stage 1 & 68.70 & 65.14 & 76.28 \\
    Stage 1 + Blur & 69.16 & 66.26 & 76.19 \\
    Stage 1 + Noise & 68.48 & 65.00 & 77.76 \\
    Stage 1 + Image swap & 67.65  & 66.36 & \textbf{69.63} \\
    \midrule
    \rowcolor{gray!20}
    \textbf{\modelName~(Ours)} & \textbf{69.87} & \textbf{67.81} & 73.27 \\
    \bottomrule[1.2pt]
    \end{tabular}
}
\label{tab:abla_hag}
\vspace{-15pt}
\end{table}

\vspace{1mm}\noindent
\textbf{Effectiveness of TopoHG.}
Here, we aim to answer two research questions: \textit{(1) Can TopoHG truly create \textbf{pseudo-break} predictions?} \textit{(2) Is the low-frequency swap method in TopoHG more effective for enhancing topological continuity compared to other data augmentation methods?}

\begin{figure}[t]
    \setlength{\abovecaptionskip}{5pt}
	\setlength{\belowcaptionskip}{-10pt}
	\centering 
	\includegraphics[width=0.49\textwidth]{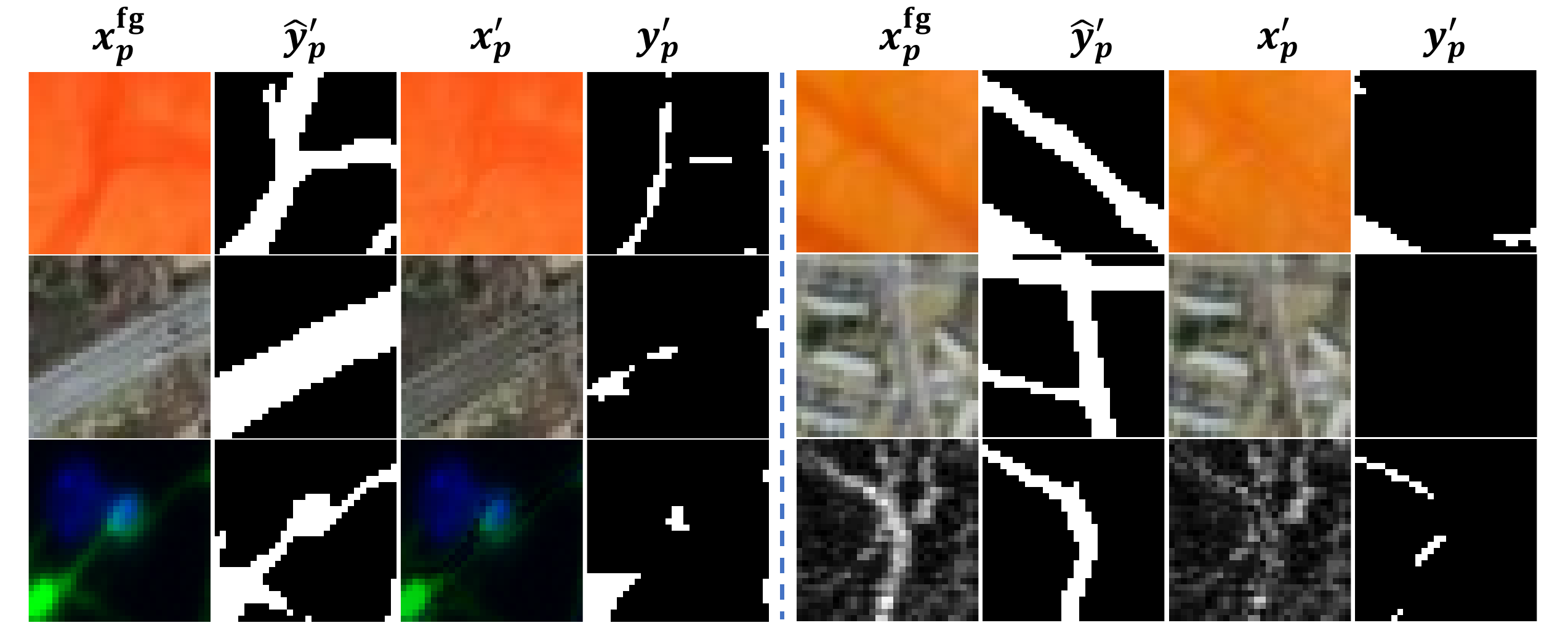}
	\caption{Visualizations of original patches $\boldsymbol x_p^{\text{fg}}$, pseudo-labels $\boldsymbol{\hat{y}}_{p}^{\prime}$, \textit{pseudo-break} patches $\boldsymbol{x}_p^{\prime}$, and weakened prediction masks $\boldsymbol{y}_p^{\prime}$ in selected regions across four scenarios.} 
	\label{fig:visual_hg}
    \vspace{-5pt}
\end{figure}

As shown in \tabref{tab:abla_main}, using only Stage 2 with TopoHG still leads to improvements in both segmentation performance and topological continuity. Compared to Stage 1 alone, adding TopoHG notably enhances continuity. To answer the first question, we provide visualizations of selected regions’ predictions before and after introducing \textit{pseudo-breaks} in \figref{fig:visual_hg}. The images generated by TopoHG significantly disrupt the model's previously confident predictions, effectively creating local \textit{pseudo-breaks}.
For the second question, \tabref{tab:abla_hag} compares various data augmentation methods, including Gaussian blur, random Gaussian noise, and foreground swap with the background in the spatial domain (image swap). The blur method creates local hard samples by removing high-frequency details, which offers some improvement. The image swap method seems to yield superior topological performance, but a sharp drop in Dice indicates that lower \textit{Betti errors} may stem from overly aggressive predictions connecting many regions incorrectly. This is likely due to consistent regularization on the background without high-frequency details, which can cause incorrect segmentation of the background as foreground. Due to the limited space, visual comparisons of different data augmentation methods are provided in Appendix.

\section{Conclusion}

In this work, we proposed the first TTA framework, i.e., \modelName, for TSS to enhance cross-domain performance by addressing discrepant topological structures and fragile topological continuity.
To improve the adaptability to topological structure, we proposed TopoMDCs to learn how to adaptively integrate various orientation information according to the topological feature of each test sample. Besides, we developed TopoHG to generate \textit{pseudo-breaks} to enhance the predictions’ topological continuity. Extensive experiments across four scenarios verified the superiority of our TopoTTA compared to other methods. 

\noindent\textbf{Acknowledgement.} This research was supported by the National Natural Science Foundation of China (Grant No. 62202027).

{
    \small
    \bibliographystyle{ieeenat_fullname}
    \bibliography{main}
}


\input{X_suppl}

\end{document}

%% file: preamble.tex
\usepackage[dvipsnames]{xcolor}


%% file: X_suppl.tex
\clearpage
\setcounter{page}{1}
\maketitlesupplementary
\appendix
\renewcommand{\thefigure}{\Alph{section}.\arabic{figure}}
\renewcommand{\thetable}{\Alph{section}.\arabic{table}} 
\setcounter{figure}{0}
\setcounter{table}{0}
\section{Overview}
This appendix is structured as follows:
\begin{itemize}
    \item In Sec.~\ref{sec:detail}, we provide more information on implementation details, including the process of TopoTTA, the implementation details in Stage 2, resizing the ground-truth label, more TopoMDCs, and training the source model.
    \item In Sec.~\ref{sec:add_exp}, we present additional experiment results, \ie additional comparison results, additional ablation study, and additional visualization results.\footnote{Code will be released at https://anonymous.4open.science/r/TopoTTA-82A0.}
\end{itemize}

\section{More Implementation Details}
\label{sec:detail}

\subsection{The Process of TopoTTA}

The overall process of TopoTTA is summarized in Algorithm~\ref{algorithm}.

\vspace{-2mm}
\begin{algorithm}[h]
\caption{The Process of TopoTTA}\label{algorithm}
\SetKwInOut{Input}{Input}\SetKwInOut{Output}{Output}
\Input{A source pre-trained TSS model $\mathcal F(\cdot;\theta)$, teacher model $\mathcal F(\cdot;\theta^{\prime})$ target domain dataset $\mathcal{D}^{t}=\{\boldsymbol x_i^{t}\}_{i=1}^{N^{t}}$, learning rates $\alpha_1$ and $\alpha_2$, number of iteration $iterations$}
\vspace{1mm}
\Output{Final prediction $\boldsymbol{\hat{y}}_{\text{out}}$}
\vspace{1mm}

\For{$\boldsymbol x \in \mathcal{D}^{t}$}{
    \textcolor{teal}{\# Stage 1: Topological structure adaptation}\\
        Define TopoMDCs by Eq.~(\ref{eq:3},\ref{eq:4},\ref{eq:5})\;
        $\mathcal{E} \gets \text{Encoder}\left(\mathcal{F}\left(\cdot;\theta\right)\right)$;~~\textcolor{teal}{\# Extract encoder from $\mathcal{F}(\cdot;\theta)$} \\
        \For{$3 \times 3~Conv$ in $\mathcal{E}$}{
            Replace $Conv$ with TopoMDCs$(\cdot; \theta; \boldsymbol{\delta})$\;}
        
        \For{$j \leftarrow 1$ \textbf{to} $iterations/2$ }{
            $\boldsymbol{\delta}  \gets  \boldsymbol{\delta}-\alpha_1\cdot\nabla_{\boldsymbol{\delta}}\mathcal L_{\text{EM}}(\mathcal{F}(\boldsymbol x;\theta; \boldsymbol{\delta}))$\;}
    \textcolor{teal}{\# Stage 2: Topological continuity refinement}\\
        \For{$j \leftarrow 1$ \textbf{to} $iterations/2$ }{
            $\boldsymbol{\hat{y}}^{\prime} \gets \mathcal{F}(\boldsymbol x;\theta^{\prime}; \boldsymbol{\delta})$\;
            Select $N_p$ points as key points\;
            \For{$p=(u_c, v_c) \leftarrow 1$ \textbf{to} $N_p$}{
                $\boldsymbol x_p^{\text{fg}}$ centered at point $p$, with a size of $s \times s$ \;
                $\boldsymbol x_p^{\text{bg,*}} \gets \mathop{\arg\min}\limits_{\boldsymbol x_p^{\text{bg}}} Confidence(\boldsymbol{\hat{y}}^{\prime,\text{bg}}_p)$;~~\textcolor{teal}{\# $\boldsymbol x_p^{\text{bg}}$ denotes the background sliding window around $\boldsymbol x_p^{\text{fg}}$} \\
                Obtain $\boldsymbol x_p^{\text{swap}}$ using low-frequency swapping by Eq.~(\ref{eq:7})\;
                Obtain \textit{pseudo-break} patch $\boldsymbol x_p^{\prime}$ by Eq.~(\ref{eq:8})\;
                }

            $\theta  \gets  \theta-\alpha_2\cdot\nabla_{\theta}\mathcal L_{\text{CE}}(\boldsymbol{\hat{y}}^{\prime},\mathcal{F}(\boldsymbol x^{\prime};\theta; \boldsymbol{\delta}))$\;}
    \textcolor{teal}{\# Prediction}\\
    $\boldsymbol{\hat{y}}_{\text{out}}  \gets  \mathcal{F}(\boldsymbol x;\theta; \boldsymbol{\delta})$\\
}
\end{algorithm}
\vspace{-2mm}

\subsection{Other Implementation Details in Stage 2}

For the Stage 2, we follow the CoTTA's settings\cite{tta_cotta}. Before inputting the image $\boldsymbol x$ into the teacher model, it undergoes four rounds of augmentations, combining random horizontal flips, vertical flips, and scaling by factors of 0.5, 1.0, 1.25, and 1.5. The average prediction from these augmented images is used as the pseudo-label. In the student model, the hard sample $\boldsymbol x^{\prime}$ undergoes a single round of the same augmentation combinations. The Baseline\textsuperscript{$\star$} in the ablation study already incorporates these settings.

\subsection{Resizing the Ground-Truth Label}

\begin{figure*}[h]
	\setlength{\abovecaptionskip}{6pt}
	\setlength{\belowcaptionskip}{-10pt}
	\centering 
	\includegraphics[width=0.7\textwidth]{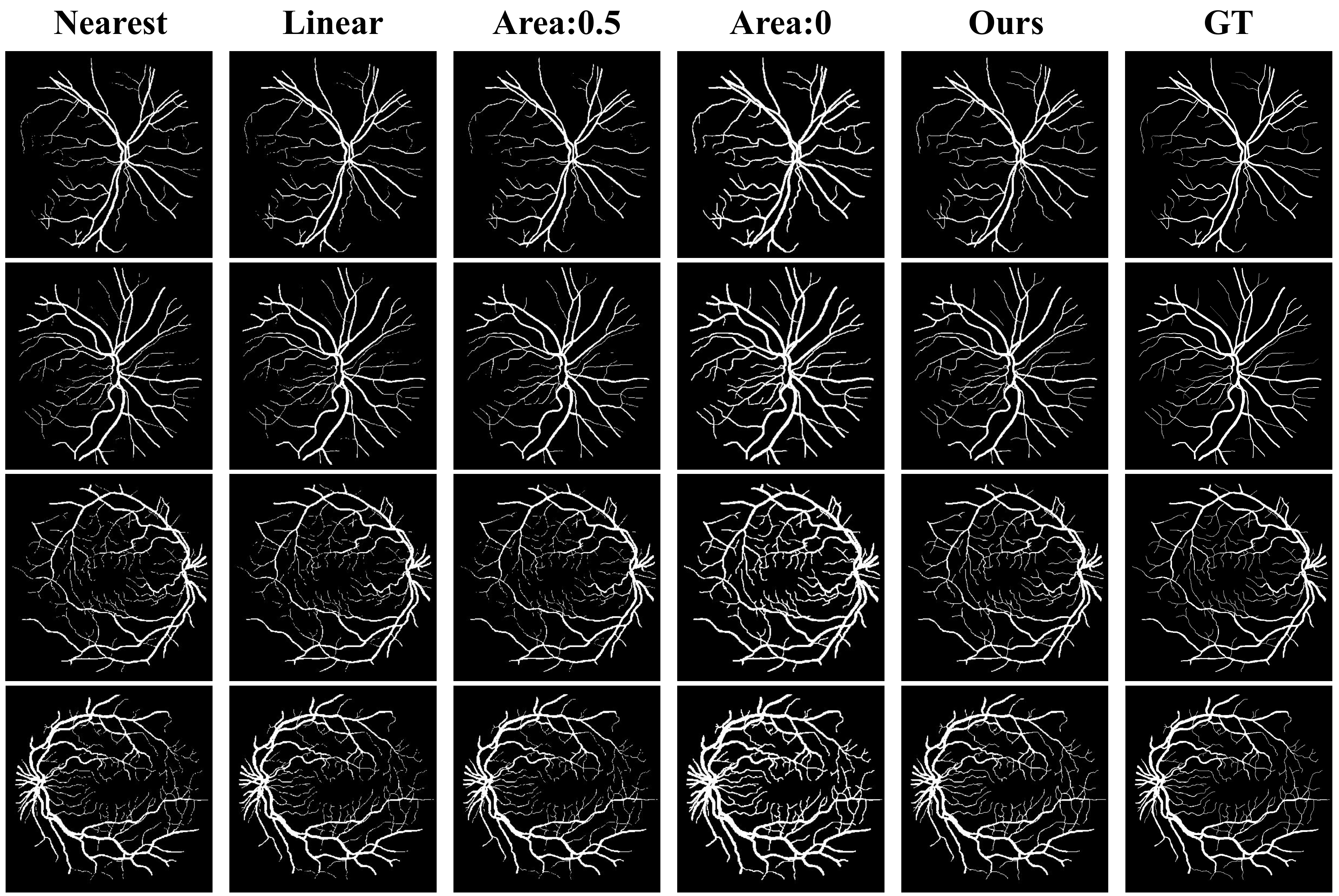}
	\caption{Visualization of the proposed resizing method compared with other commonly used approaches. Area:0.5 represents the image binarized with a threshold of 0.5, Area:0 uses a threshold of 0, and GT is short for ground-truth labels.} 
	\label{fig:resize} 
\end{figure*}

In the manuscript, we discuss resizing the images in datasets from three scenarios to $384\times384$. However, conventional nearest or linear interpolation methods often cause breakage in the thin annotated regions of the ground-truth labels, negatively affecting both training and topological metric calculations. To address this issue, we propose a novel resizing method. Specifically, the images are first resized using the area-based interpolation method (available in the OpenCV library). Binarization is then applied with thresholds of 0.5 and 0, as shown in the third and fourth columns of Fig.~\ref{fig:resize}. Next, the skeleton map is extracted from the binarized result using a threshold of 0. The skeleton map and the binarized image obtained with a threshold of 0.5 are combined using a pixel-wise OR operation to produce the final resized ground-truth labels, as shown in the fifth column of Fig.~\ref{fig:resize}. This resizing method produces results that closely resemble the original ground-truth labels and effectively preserves topological consistency.

\vspace{-3mm}

\subsection{More Topology-Meta Difference Convolutions}

In the manuscript, we present the formulation for calculating $\mathcal{C}_1$. Here, we provide the formulations for all other directions of TopoMDCs,
\begin{equation}
    \begin{split}
    \mathcal C_i(r_x, r_y) &= \mathcal C_c(r_x, r_y) - \!\!\!\!\!\!\!\!\!\!\sum_{(\Delta r_x,\Delta r_y) \in \mathcal R_i} \!\!\!\!\!\!\!\!\!\!w(\Delta r_x, \Delta r_y) \cdot \boldsymbol x_{\text{in}}(r_x, r_y)  \\
     &+\!\!\!\!\!\!\!\!\!\!\!\!\!\!\!\!\!\!\!\!\!\!\!\!\sum_{(\Delta r_x,\Delta r_y) \in \mathcal R_i,~(\Delta b_x,\Delta b_y) \in \mathcal B_i} \!\!\!\!\!\!\!\!\!\!\!\!\!\!\!\!\!\!\!\!\!\!\!\!\!w(\Delta r_x, \Delta r_y) \cdot \boldsymbol x_{\text{in}}(r_x - \Delta b_x, r_y - \Delta b_y),
    \end{split}
\end{equation}
\begin{equation}
    \begin{split}
    &\mathcal R_1 = \{(-1,-1),(-1,0),(0,-1)\},~~\mathcal B_1 = \{(-1,-1)\};~~\mathcal R_2 = \{(0,-1),(-1,-1),(1,-1)\},~~\mathcal B_2 = \{(0,-1)\};\\
    &\mathcal R_3 = \{(1,-1),(0,-1),(1,0)\},~~~~~~~~\mathcal B_3 = \{(1,-1)\};~~~~~\mathcal R_4 = \{(-1,0),(-1,-1),(-1,1)\},~~\mathcal B_4 = \{(-1,0)\}; \\
    &\mathcal R_5 = \{(1,0),(1,1),(1,-1)\},~~~~~~~~~~~\mathcal B_5 = \{(1,0)\};~~~~~~~~\mathcal R_6 = \{(-1,1),(-1,0),(0,1)\},~~~~~~~~\mathcal B_6 = \{(-1,1)\}; \\
    &\mathcal R_7 = \{(0,1),(1,1),(-1,1)\},~~~~~~~~~~~\mathcal B_7 = \{(0,1)\};~~~~~~~~\mathcal R_8 = \{(1,1),(0,1),(1,0)\},~~~~~~~~~~~~~~\mathcal B_8 = \{(1,1)\}. \\
    \end{split}
\end{equation}

\vspace{-3mm}
\begin{figure*}[h]
	\setlength{\abovecaptionskip}{6pt}
	\setlength{\belowcaptionskip}{-10pt}
	\centering 
	\includegraphics[width=0.45\textwidth]{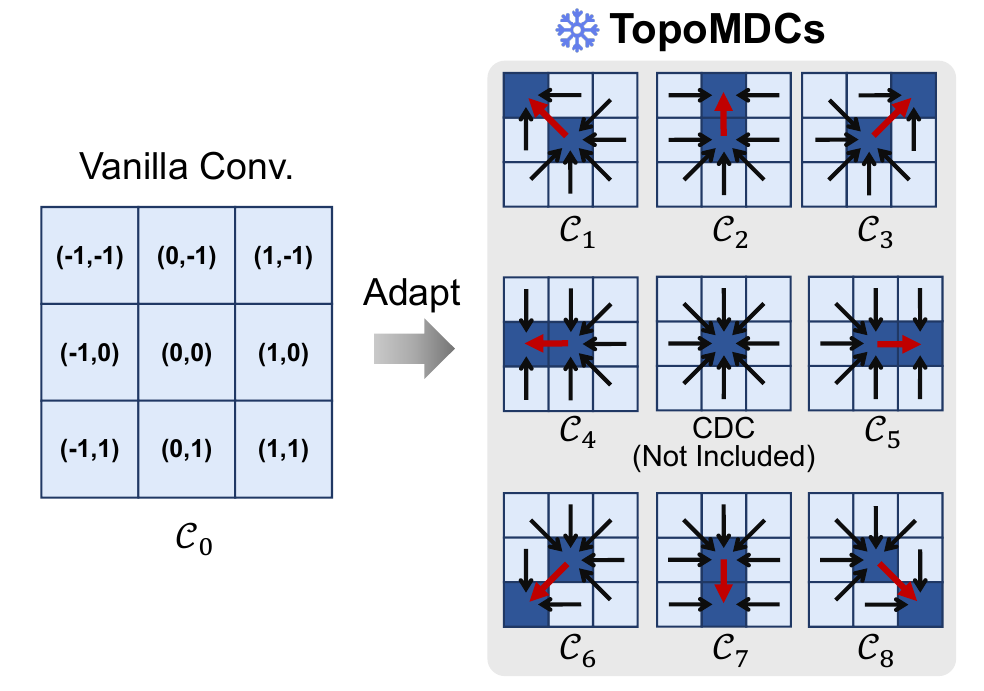}
	\caption{Different Topology-Meta Difference Convolutions, which inherit parameters from the source domain model without adding additional parameters to convolution layers.} 
	\label{fig:topomdc} 
\end{figure*}

\subsection{Training the Source Model}

The Adam optimizer with a learning rate of $5\times10^{-4}$ is used across all four scenarios, with the batch size set to 4. The maximum number of epochs is set to 100, 50, 100, and 60 for the four scenarios, respectively. The model with the best performance on the source domain test set is selected for testing. Training is conducted using a combination of Dice and binary cross entropy (BCE) loss functions. During training, random horizontal and vertical flips are applied as data augmentations.

\section{Additional Experimental Results}
\label{sec:add_exp}
\subsection{Additional Comparison Results}

Table~\ref{tab:sup1} presents the detailed experimental results of TopoTTA using CS2Net as the baseline network across four scenarios. The conclusions are consistent with those drawn when UNet is used as the baseline network: TopoTTA delivers significant improvements in both segmentation performance and topological connectivity in most scenarios. Table~\ref{tab:sup2} shows the performance of a TopoTTA variant applied to DSCNet. In this variant, Stage 1 is omitted, and only Stage 2 is used for parameter updates. This adjustment is necessary because DSCNet already incorporates deformable convolutional kernels, which limit the compatibility of TopoMDCs. As shown in Table~\ref{tab:sup2}, even with this simplified version, TopoTTA still outperforms most comparison methods across the majority of scenarios. These additional experiments further validate the broad applicability and effectiveness of TopoTTA, demonstrating its ability to efficiently adapt to different CNN-based models. Table~\ref{tab:sup3} presents the paired t-test results of clDice. The results indicate that the \textit{p}-value $\textless 0.05$ almost across all datasets, inferring TopoTTA's improvement is significant. 

\vspace{-10pt}
\begin{table*}[h]
	\caption{Cross-domain testing results obtained using CS2Net as baseline network in four different scenarios, i.e., retinal vessel segmentation, road extraction, microscopic neuronal segmentation, and retinal OCT-angiography vessel segmentation. Source Only: Trained on the source, and tested on the target domain directly. The best and second-best results in each column are highlighted in \textbf{bold} and \underline{underline}.}
        \vspace{-5pt}
	\setlength{\tabcolsep}{5pt}
        \renewcommand\arraystretch{0.95}
	\resizebox*{1.0 \linewidth}{!}{
		\begin{tabular}{lccccccccccccccc}
			\toprule[1.2pt]
			\multirow{2}{*}{\textbf{Method}} & \multicolumn{3}{c}{\textbf{DRIVE $\rightarrow$ CHASE}} &  \multicolumn{3}{c}{\textbf{DRIVE $\rightarrow$ STARE}} &  \multicolumn{3}{c}{\textbf{CHASE $\rightarrow$ DRIVE}} &  \multicolumn{3}{c}{\textbf{CHASE $\rightarrow$ STARE}} &  \multicolumn{3}{c}{\textbf{DeepGlobe $\rightarrow$ MR}} \\
			\cmidrule(lr){2-4}\cmidrule(lr){5-7}\cmidrule(lr){8-10}\cmidrule(lr){11-13}\cmidrule(lr){14-16}
			& Dice (\%) $\uparrow$ & clDice (\%) $\uparrow$ & $\beta$ $\downarrow$ & Dice (\%) $\uparrow$ & clDice (\%) $\uparrow$ & $\beta$ $\downarrow$ & Dice (\%) $\uparrow$ & clDice (\%) $\uparrow$ & $\beta$ $\downarrow$ & Dice $\uparrow$ & clDice (\%) $\uparrow$ & $\beta$ $\downarrow$ & Dice (\%) $\uparrow$ & clDice (\%) $\uparrow$ & $\beta$ $\downarrow$ \\
                \midrule[0.7pt]

			Source Only & 22.58  & 19.03  &  43.00  & 37.27  & 33.94  & 112.35  &  61.78  & 62.33  &  95.35 &  48.46  &  46.73  &  107.95  &  45.70 &  55.85 & 79.65 \\
			TENT \cite{tta_tent} & 64.01  & 65.22  &  33.50  & 60.93  &  \underline{55.65}  & 107.80  &  67.53  & 67.14  &  88.65 &  60.88  &  56.91  &  107.75  &  42.41 &  52.49 & 77.88 \\
                CoTTA \cite{tta_cotta} & \underline{67.04}  & \underline{69.08}  &  \underline{28.50}  & \underline{61.40}  &  55.42  & \underline{102.10}  &  \underline{67.95}  & \underline{67.44}  &  \underline{83.90} &  \underline{62.08}  &  \underline{58.00}  &  \underline{102.35}  &  \textbf{48.88} & \underline{58.55}  & \underline{74.65}\\
			SAR \cite{tta_SAR} & 63.72  & 64.86  &  33.00  & 60.77  & 55.52 & 109.15  &  67.17  & 66.83  &  88.95 & 60.81  &  56.86  &  107.10  &  43.69 & 55.02  & 77.11 \\
                DIGA \cite{tta_diga} & 63.87  & 64.49  & 34.13  & 59.47  &  54.34  & 108.40  & 67.78  & 67.34  & 87.15 & 61.67 & 57.45  & 107.70 & 43.71  &  55.02 & 76.96 \\
			DomainAdaptor \cite{tta_domainadaptor} & 58.39  & 57.34  & 41.38  & 54.76  &  49.29  & 111.10  & 65.21  & 65.30  & 89.80 & 57.11  &  54.00  &  112.40  & 43.55  &  56.09 & 77.37 \\
			MedBN \cite{tta_medbn} & 58.10  & 58.38  &  34.63  & 57.40  &  51.92  & 116.40  &  65.21  & 63.01  &  90.95 & 61.76  &  56.85  &  110.35  & 43.42  &  53.49 & 77.53 \\
			VPTTA \cite{tta_vptta} & 62.14  & 62.88  &  35.26  & 60.59  &  54.93  & 108.00  &  66.97  & 66.66  &  88.95 & 60.30  &  56.46  &  107.50  &  43.89 &  55.31 & 77.41  \\
                \midrule
                \rowcolor{gray!20}
			\textbf{\modelName~(Ours)} & \textbf{68.27}  & \textbf{72.99}  &  \textbf{21.88} &  \textbf{62.00}  & \textbf{57.46}  &  \textbf{91.60}  & \textbf{71.12}  &  \textbf{69.74}  &  \textbf{77.50}  &  \textbf{65.09}  &  \textbf{59.91}  &  \textbf{96.20}  &  \underline{48.47} &  \textbf{63.09} & \textbf{73.55} \\
                \midrule[1.0pt]
                \multirow{2}{*}{\textbf{Method}} & \multicolumn{3}{c}{\textbf{DeepGlobe $\rightarrow$ CNDS}} &  \multicolumn{3}{c}{\textbf{Neub1 $\rightarrow$ Neub2}} &  \multicolumn{3}{c}{\textbf{Neub2 $\rightarrow$ Neub1}} &  \multicolumn{3}{c}{\textbf{ROSE $\rightarrow$ OCTA500}} &  \multicolumn{3}{c}{\textbf{OCTA500 $\rightarrow$ ROSE}} \\
			\cmidrule(lr){2-4}\cmidrule(lr){5-7}\cmidrule(lr){8-10}\cmidrule(lr){11-13}\cmidrule(lr){14-16}
			& Dice (\%) $\uparrow$ & clDice (\%) $\uparrow$ & $\beta$ $\downarrow$ & Dice (\%) $\uparrow$ & clDice (\%) $\uparrow$ & $\beta$ $\downarrow$ & Dice (\%) $\uparrow$ & clDice (\%) $\uparrow$ & $\beta$ $\downarrow$ & Dice (\%) $\uparrow$ & clDice (\%) $\uparrow$ & $\beta$ $\downarrow$ & Dice (\%) $\uparrow$ & clDice (\%) $\uparrow$ & $\beta$ $\downarrow$ \\
                \midrule[0.7pt]
            
			Source Only  &  \underline{84.28} & 91.57  & 10.95 & 22.00  & /  &  \underline{7.56}  & 61.76  &  72.72  & 7.31  &  46.94  & 53.88  &  61.00 &  72.36  & 75.60  &  18.33\\
			TENT \cite{tta_tent} &  76.55 &  89.29 & 7.50 & 62.26  & 70.52  &  7.95  & 62.64  &  74.01  & 7.03  &  66.76  & 72.10  &  64.44 &  73.58 &  77.60 &  16.23\\
                CoTTA \cite{tta_cotta} &  75.43 &  89.72 & \underline{6.98} & \underline{63.88}  & \underline{72.98}  &  8.22  & \underline{63.66}  &  \underline{75.87}  & \underline{5.78}  &  \textbf{68.82}  & \underline{75.01}  &  \underline{52.48} &  72.07 & 76.34  & 16.89 \\
			SAR \cite{tta_SAR} & 77.02  & 89.23  & 7.77 & 61.55  & 69.81  & 8.33  & 61.86  & 73.17 & 7.22  &  65.99  & 72.10  &  63.76 & 73.67 &  77.68 &  16.00\\
                DIGA \cite{tta_diga} & 79.72  & 89.70  & 7.73  & 61.31  &  70.40  & 8.05  & 63.57  & 73.90  & 7.31 & 66.99  & 73.02  & 65.60 & 71.09  &  76.38 & 16.66 \\
			DomainAdaptor \cite{tta_domainadaptor} &  78.81 &  90.94 & 7.61 & 53.99  & 63.87  & 10.67  & 63.12  &  74.19  & 6.97  & 63.12  & 69.60  & 72.90 & 71.45 &  75.88 & 16.56\\
			MedBN \cite{tta_medbn} & 81.22  & \underline{92.16}  & 8.08 & 56.08  & 63.72  &  11.50  & 63.56  &  72.80 & 9.84  &  63.97  & 69.11  &  78.92 & \underline{75.55} & \underline{78.28}  & \textbf{8.11} \\
			VPTTA \cite{tta_vptta} & 77.32  &  89.52 & 7.68 & 60.87  & 69.56  &  9.28  & 62.21  &  73.47  & 7.32  &  65.32  & 71.54  &  65.28 & 73.34 &  77.41  & 15.77 \\
                \midrule
                \rowcolor{gray!20}
			\textbf{\modelName~(Ours)} & \textbf{87.89}  &  \textbf{96.05} & \textbf{4.24} & \textbf{66.73}  & \textbf{75.81}  &  \textbf{6.72}  &  \textbf{64.00}  &  \textbf{76.60}  & \textbf{5.31}  &  \underline{67.62}  & \textbf{76.65}  &  \textbf{45.92}  & \textbf{75.57} &  \textbf{78.72} &  \underline{15.67} \\
			\bottomrule[1.2pt]
		\end{tabular}
	}
    \vspace*{-5pt}
    \label{tab:sup1}
\end{table*}

\vspace{-15pt}
\begin{table*}[h]
	\caption{Cross-domain testing results obtained using DSCNet as baseline network in four different scenarios, i.e., retinal vessel segmentation, road extraction, microscopic neuronal segmentation, and retinal OCT-angiography vessel segmentation. Source Only: Trained on the source, and tested on the target domain directly. The best and second-best results in each column are highlighted in \textbf{bold} and \underline{underline}.}
        \vspace{-5pt}
	\setlength{\tabcolsep}{5pt}
        \renewcommand\arraystretch{0.95}
	\resizebox*{1.0 \linewidth}{!}{
		\begin{tabular}{lccccccccccccccc}
			\toprule[1.2pt]
			\multirow{2}{*}{\textbf{Method}} & \multicolumn{3}{c}{\textbf{DRIVE $\rightarrow$ CHASE}} &  \multicolumn{3}{c}{\textbf{DRIVE $\rightarrow$ STARE}} &  \multicolumn{3}{c}{\textbf{CHASE $\rightarrow$ DRIVE}} &  \multicolumn{3}{c}{\textbf{CHASE $\rightarrow$ STARE}} &  \multicolumn{3}{c}{\textbf{DeepGlobe $\rightarrow$ MR}} \\
			\cmidrule(lr){2-4}\cmidrule(lr){5-7}\cmidrule(lr){8-10}\cmidrule(lr){11-13}\cmidrule(lr){14-16}
			& Dice (\%) $\uparrow$ & clDice (\%) $\uparrow$ & $\beta$ $\downarrow$ & Dice (\%) $\uparrow$ & clDice (\%) $\uparrow$ & $\beta$ $\downarrow$ & Dice (\%) $\uparrow$ & clDice (\%) $\uparrow$ & $\beta$ $\downarrow$ & Dice $\uparrow$ & clDice (\%) $\uparrow$ & $\beta$ $\downarrow$ & Dice (\%) $\uparrow$ & clDice (\%) $\uparrow$ & $\beta$ $\downarrow$ \\
                \midrule[0.7pt]
                
			Source Only & 23.16 & 18.98 & 37.25 & 42.46 & 36.89 & 102.35 & 58.58 & 53.95 & 94.60 & 44.19 & / & 106.90 & 42.71 & 51.87 & 81.28 \\
			TENT \cite{tta_tent} & 64.17 & 65.42 & 22.38 & 56.87 & 50.50 & 105.90 & 66.75 & 62.32 & 83.80 & 62.18 & 55.28 & 102.85 & 40.30 & 47.69 & 79.94 \\
                CoTTA \cite{tta_cotta} & \underline{67.14} & \underline{69.83} & 25.13 & 57.69 & 51.30 & 103.50 & \underline{68.80} & \underline{65.04} & \underline{79.85} & \underline{63.96} & \underline{57.54} & \underline{99.70} & \underline{45.13} & 54.00 & \textbf{77.68} \\
			SAR \cite{tta_SAR} & 64.08 & 65.30 & 23.25 & 56.98 & 50.67 & 105.75 & 66.57 & 62.12 & 84.30 & 62.18 & 55.32 & 102.90 & 42.61 & 51.79 & 80.48 \\
                DIGA \cite{tta_diga} & 63.87 & 66.58 & 24.38 & \underline{57.82} & \underline{51.76} & 105.75 & 65.58 & 60.95 & 85.15 & 61.43 & 54.50 & 104.35 & 42.83 & \underline{54.67} & 78.63 \\
			DomainAdaptor \cite{tta_domainadaptor} & 60.42 & 43.22 & 105.90 & 49.74 & 43.22 & 105.90 & 65.60 & 61.44 & 85.25 & 61.99 & 54.02 & 103.50 & 43.14 & 53.12 & 79.24 \\
			MedBN \cite{tta_medbn} & 58.36 & 58.80 & 29.63 & 54.57 & 49.68 & \underline{96.30} & 66.15 & 64.52 & 81.55 & 61.50 & 57.72 & 102.50 & 41.71 & 49.59 & 83.96 \\
			VPTTA \cite{tta_vptta} & 63.22 & 64.24 & \underline{21.13} & 56.44 & 49.85 & 104.50 & 66.50 & 62.11 & 83.55 & 62.29 & 55.18 & 103.35 & 42.41 & 51.15 & 79.77  \\
                \midrule
                \rowcolor{gray!20}
			\textbf{\modelName~(Stage2 only)} & \textbf{67.32} & \textbf{71.89} & \textbf{19.13} & \textbf{60.24} & \textbf{53.06} & \textbf{94.25} & \textbf{70.35} & \textbf{66.88} & \textbf{79.45} & \textbf{64.63} & \textbf{58.05} & \textbf{95.35} & \textbf{47.17} & \textbf{61.55} & \underline{77.82} \\
                \midrule[1.0pt]
                \multirow{2}{*}{\textbf{Method}} & \multicolumn{3}{c}{\textbf{DeepGlobe $\rightarrow$ CNDS}} &  \multicolumn{3}{c}{\textbf{Neub1 $\rightarrow$ Neub2}} &  \multicolumn{3}{c}{\textbf{Neub2 $\rightarrow$ Neub1}} &  \multicolumn{3}{c}{\textbf{ROSE $\rightarrow$ OCTA500}} &  \multicolumn{3}{c}{\textbf{OCTA500 $\rightarrow$ ROSE}} \\
			\cmidrule(lr){2-4}\cmidrule(lr){5-7}\cmidrule(lr){8-10}\cmidrule(lr){11-13}\cmidrule(lr){14-16}
			& Dice (\%) $\uparrow$ & clDice (\%) $\uparrow$ & $\beta$ $\downarrow$ & Dice (\%) $\uparrow$ & clDice (\%) $\uparrow$ & $\beta$ $\downarrow$ & Dice (\%) $\uparrow$ & clDice (\%) $\uparrow$ & $\beta$ $\downarrow$ & Dice (\%) $\uparrow$ & clDice (\%) $\uparrow$ & $\beta$ $\downarrow$ & Dice (\%) $\uparrow$ & clDice (\%) $\uparrow$ & $\beta$ $\downarrow$ \\
                \midrule[0.7pt]
            
			Source Only  &  81.80 & 88.67 & 14.73 & 3.94 & / & 9.44 & 58.66 & 58.97 & 7.88 & 58.76 & 65.87 & 61.44 & 71.22 & 74.31 & 17.66 \\
			TENT \cite{tta_tent} &  80.11 & 92.78 & 7.27 & 54.75 & 63.51 & 7.83 & 58.52 & 58.05 & 7.47 & 67.36 & 74.11 & 48.66 & 72.60 & 76.67 & \textbf{15.88} \\
                CoTTA \cite{tta_cotta} &  80.26 & \underline{93.32} & \underline{6.82} & \underline{56.49} & \underline{64.58} & 10.65 & \textbf{62.93} & \underline{65.72} & 7.00 & \textbf{68.81} & \underline{73.99} & \underline{46.54} & 71.73 & 76.65 & 17.78 \\
			SAR \cite{tta_SAR} & 81.15 & 92.67 & 7.80 & 55.04 & 63.77 & 8.28 & 58.55 & 58.16 & 7.38 & 66.86 & 73.24 & 50.6 & 72.75 & \underline{76.82} & 16.11 \\
                DIGA \cite{tta_diga} & \underline{83.76} & 92.69 & 8.93 & 55.82 & 64.33 & 9.34 & 58.77 & 60.72 & \underline{6.91} & 67.35 & 73.74 & 52.04 & 69.27 & 75.19 & 17.89 \\
			DomainAdaptor \cite{tta_domainadaptor} &  78.01 & 90.82 & 7.37 & 37.31 & 46.80 & \textbf{6.84} & 58.57 & 58.53 & 7.28 & 65.49 & 72.36 & 54.78 & 72.38 & 76.36 & 17.00 \\
			MedBN \cite{tta_medbn} & 80.35 & 92.16 & 11.75 & 51.54 & 59.16 & 7.84 & 58.58 & 58.55 & 8.19 & 57.41 & 63.26 & 60.06 & 70.76 & 73.86 & 18.00 \\
			VPTTA \cite{tta_vptta} & 79.68 & 92.07 & 7.40 & 53.16 & 61.53 & \underline{7.44} & 58.55 & 58.09 & 7.47 & 66.77 & 73.23 & 50.70 & \underline{72.81} & 76.76 & \underline{16.00} \\
                \midrule
                \rowcolor{gray!20}
			\textbf{\modelName~(Stage2 only)} & \textbf{88.63} & \textbf{96.62} & \textbf{4.77} & \textbf{57.45} & \textbf{66.70} & 8.27 & \underline{62.88} & \textbf{71.28} & \textbf{5.72} & \underline{68.50} & \textbf{75.55} & \textbf{37.32} & \textbf{74.03} & \textbf{77.51} & 17.33 \\
			\bottomrule[1.2pt]
		\end{tabular}
	}
    \vspace*{-5pt}
    \label{tab:sup2}
\end{table*}

\vspace{-15pt}
\begin{table*}[h]
	\centering
    \caption{The paired t-test results of clDice using UNet as baseline network across ten datasets. Source Only: Trained on the source, and tested on the target domain directly.}
        \vspace{-5pt}
	\setlength{\tabcolsep}{5pt}
        \renewcommand\arraystretch{0.95}
	\resizebox*{0.85 \linewidth}{!}{
		\begin{tabular}{lccccc}
			\toprule[1.2pt]
			\multirow{2}{*}{\textbf{Method}} & \multicolumn{5}{c}{\textit{p} value} \\
            \cmidrule(lr){2-6}
            & {\textbf{DRIVE $\rightarrow$ CHASE}} &  {\textbf{DRIVE $\rightarrow$ STARE}} &  {\textbf{CHASE $\rightarrow$ DRIVE}} &  {\textbf{CHASE $\rightarrow$ STARE}} &  {\textbf{DeepGlobe $\rightarrow$ MR}} \\
                \midrule[0.7pt]
			Source Only & $\textless 0.001$ & $\textless 0.001$ & $\textless 0.001$ & $\textless 0.001$ & $\textless 0.001$ \\
			TENT \cite{tta_tent} & 0.0011 & 0.0045 & $\textless 0.001$ & $\textless 0.001$ & $\textless 0.001$ \\
                CoTTA \cite{tta_cotta} & 0.0128 & $\textless 0.001$ & $\textless 0.001$ & 0.0033 & 0.001 \\
			SAR \cite{tta_SAR} & 0.001 & 0.0085 & $\textless 0.001$ & $\textless 0.001$ & $\textless 0.001$ \\
                DIGA \cite{tta_diga} & $\textless 0.001$ & 0.057 & $\textless 0.001$ & 0.027 & $\textless 0.001$ \\
			DomainAdaptor \cite{tta_domainadaptor} & 0.0011 & $\textless 0.001$ & $\textless 0.001$ & $\textless 0.001$ & $\textless 0.001$ \\
			MedBN \cite{tta_medbn} & $\textless 0.001$ & $\textless 0.001$ & $\textless 0.001$ & $\textless 0.001$ & $\textless 0.001$ \\
			VPTTA \cite{tta_vptta} & 0.0016 & 0.0032 & $\textless 0.001$ & $\textless 0.001$ & $\textless 0.001$  \\
                \midrule[1.0pt]
            \multirow{2}{*}{\textbf{Method}} & \multicolumn{5}{c}{\textit{p} value} \\
            \cmidrule(lr){2-6}
                & {\textbf{DeepGlobe $\rightarrow$ CNDS}} & {\textbf{Neub1 $\rightarrow$ Neub2}} & {\textbf{Neub2 $\rightarrow$ Neub1}} &  {\textbf{ROSE $\rightarrow$ OCTA500}} & {\textbf{OCTA500 $\rightarrow$ ROSE}} \\
                \midrule[0.7pt]
            
			Source Only & $\textless 0.001$ & $\textless 0.001$ & $\textless 0.001$ & $\textless 0.001$ & $\textless 0.001$ \\
			TENT \cite{tta_tent} & $\textless 0.001$ & 0.0036 & $\textless 0.001$ & $\textless 0.001$ & $\textless 0.001$ \\
                CoTTA \cite{tta_cotta} & $\textless 0.001$ & 0.0146 & 0.0511 & $\textless 0.001$ & 0.0067 \\
			SAR \cite{tta_SAR} & $\textless 0.001$ & 0.0053 & $\textless 0.001$ & $\textless 0.001$ & $\textless 0.001$ \\
                DIGA \cite{tta_diga} & $\textless 0.001$ & 0.0325 & 0.037 & $\textless 0.001$ & $\textless 0.001$ \\
			DomainAdaptor \cite{tta_domainadaptor} & $\textless 0.001$ & 0.0015 & $\textless 0.001$ & $\textless 0.001$ & $\textless 0.001$ \\
			MedBN \cite{tta_medbn} & $\textless 0.001$ & 0.0073 & $\textless 0.001$ & $\textless 0.001$ & 0.0013 \\
			VPTTA \cite{tta_vptta} & $\textless 0.001$ & 0.0128 & $\textless 0.001$ & $\textless 0.001$ & $\textless 0.001$ \\
			\bottomrule[1.2pt]
		\end{tabular}
	}
    \vspace*{-5pt}
    \label{tab:sup3}
\end{table*}

\subsection{Additional Ablation Study}
\label{sec:add_abla}
Due to the space limitation of our manuscript, we present a comprehensive ablation study in this section.

\vspace*{5pt}
\noindent\textbf{Impact of Different Iterations.}

In the manuscript, we set the number of iterations to six. Here, we explore the impact of different iteration counts. Table~\ref{tab:sup4} shows the performance of TopoTTA and the competing methods under varying iteration settings. DIGA, which lacks a backforward process, is excluded from this comparison. TopoTTA consistently achieves the best performance across all iteration counts. For competing methods that update only BN parameters or external parameters, performance remains stable and shows strong robustness to iteration changes. Similar to TopoTTA, CoTTA demonstrates continuous performance improvement as the number of iterations increases. However, excessive iterations result in prolonged inference times. To balance performance and efficiency, we select six as the optimal number of iterations.

\begin{table*}[h]
	\caption{Average testing results across four scenarios under different iteration counts. The best and second-best results in each column are highlighted in \textbf{bold} and \underline{underline}, respectively.}
        \vspace{-5pt}
	\setlength{\tabcolsep}{5pt}
        \renewcommand\arraystretch{1.1}
	\resizebox*{1.0 \linewidth}{!}{
		\begin{tabular}{lccccccccccccccc}
			\toprule[1.2pt]
			\multirow{2}{*}{\textbf{Method}} & \multicolumn{3}{c}{\textbf{Iterations = 2}} &  \multicolumn{3}{c}{\textbf{Iterations = 4}} &  \multicolumn{3}{c}{\textbf{Iterations = 6}} &  \multicolumn{3}{c}{\textbf{Iterations = 8}} &  \multicolumn{3}{c}{\textbf{Iterations = 10}} \\
			\cmidrule(lr){2-4}\cmidrule(lr){5-7}\cmidrule(lr){8-10}\cmidrule(lr){11-13}\cmidrule(lr){14-16}
			& Dice (\%) $\uparrow$ & clDice (\%) $\uparrow$ & $\beta$ $\downarrow$ & Dice (\%) $\uparrow$ & clDice (\%) $\uparrow$ & $\beta$ $\downarrow$ & Dice (\%) $\uparrow$ & clDice (\%) $\uparrow$ & $\beta$ $\downarrow$ & Dice $\uparrow$ & clDice (\%) $\uparrow$ & $\beta$ $\downarrow$ & Dice (\%) $\uparrow$ & clDice (\%) $\uparrow$ & $\beta$ $\downarrow$ \\
                \midrule[0.7pt]
			TENT \cite{tta_tent} & 63.94 & 67.44 & 49.44 & 63.88 & 67.38 & 49.36 & 63.81 & 67.29 & 49.42 & 63.73 & 67.19 & 49.37 & 63.62 & 67.07 & 49.34 \\
                CoTTA \cite{tta_cotta} & \underline{64.99} & \underline{68.92} & \underline{46.82} & \underline{65.33} & \underline{69.11} & \underline{47.31} & \underline{65.49} & \underline{69.34} & \underline{47.30} & \underline{66.62} & \underline{70.45} & \underline{47.06} & \underline{67.55} & \underline{71.46} & \underline{46.95}\\
			SAR \cite{tta_SAR} & 63.97 & 67.48 & 49.38 & 63.96 & 67.47 & 49.46 & 63.97 & 67.49 & 49.32 & 63.95 & 67.45 & 49.52 & 63.93 & 67.42 & 49.55 \\
			DomainAdaptor \cite{tta_domainadaptor} & 63.41 & 66.53 & 49.98 & 63.42 & 66.53 & 49.98 & 63.41 & 66.53 & 49.98 & 63.41 & 66.53 & 49.98 & 63.41 & 66.53 & 49.98\\
			MedBN \cite{tta_medbn} & 56.21 & 53.53 & 83.75 & 56.57 & 54.42 & 82.22 & 56.65 & 54.90 & 80.05 & 56.61 & 54.88 & 79.78 & 56.73 & 54.97 & 76.75\\
			VPTTA \cite{tta_vptta} & 63.98 & 67.48 & 49.75 & 63.98 & 67.48 & 49.79 & 63.98 & 67.48 & 49.76 & 63.99 & 67.49 & 49.76 & 63.99 & 67.49 & 49.76\\
                \midrule
                \rowcolor{gray!20}
			\textbf{\modelName~(Ours)} & \textbf{67.72} & \textbf{71.51} & \textbf{43.90} & \textbf{68.48} & \textbf{73.22} & \textbf{42.38} & \textbf{69.44} & \textbf{74.00} & \textbf{43.01} & \textbf{69.31} & \textbf{74.33} & \textbf{42.93} & \textbf{69.32} & \textbf{74.62} & \textbf{42.29}\\
			\bottomrule[1.2pt]
		\end{tabular}
	}
    \vspace*{-5pt}
    \label{tab:sup4}
\end{table*}

\vspace*{5pt}
\noindent\textbf{Impact of Different Values of Hyperparameters.}

\begin{figure}[h]
	\setlength{\abovecaptionskip}{6pt}
	\setlength{\belowcaptionskip}{-10pt}
	\centering 
	\includegraphics[width=0.85\textwidth]{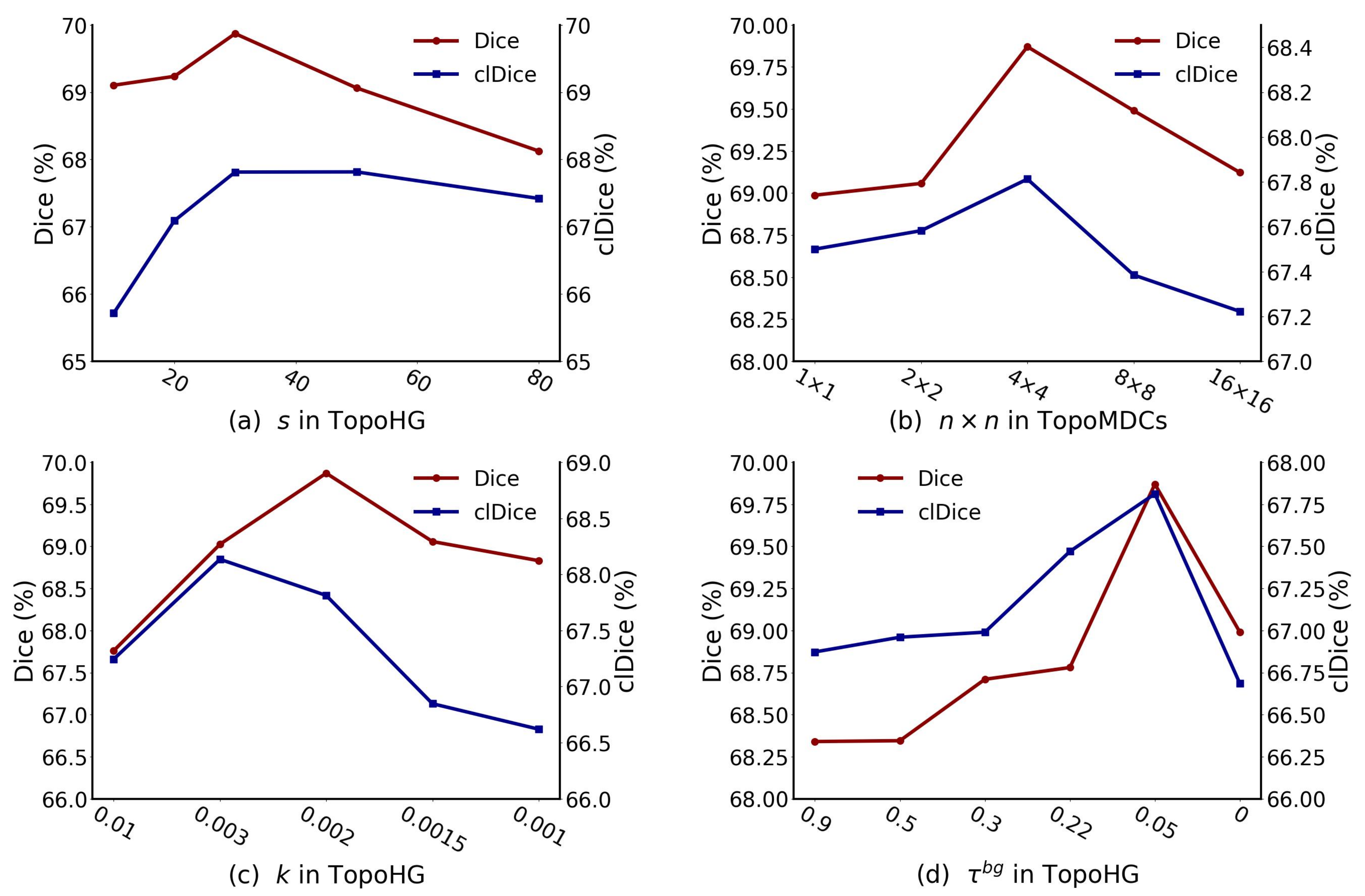}
	\caption{Performance of TopoTTA with different hyperparameter conditions.} 
	\label{fig:hyper} 
\end{figure}

To verify the sensitivity of our method to hyperparameters, we test the network performance under different hyperparameter conditions. As shown in Fig.~\ref{fig:hyper}, we analyze the $s$ (modification window size in TopoHG), $n \times n$ (number of TopoMDCs regions), $k$ (coefficient for selecting the number of key points) and $\tau^{\text{bg}}$(upper limit of the foreground pixel ratio). From Fig.~\ref{fig:hyper}(a), we can observe Dice initially rises and then drops sharply, while clDice stabilizes after reaching a peak. This is because small modification areas are insufficient to create \textit{pseudo-breaks}, whereas excessively large areas can result in falsely high continuity due to overly aggressive predictions. As shown in Fig.~\ref{fig:hyper}(b), both Dice and clDice achieve their highest values when $4 \times 4$ is used. When $n$ is too small, TopoMDCs struggle to capture the varied topological features across regions, leading to lower performance. Conversely, when $n$ is too large, the increased number of learnable $\boldsymbol{\delta}$ parameters makes learning more difficult, also reducing performance. From Fig.\ref{fig:hyper}(c), we select 0.002 as the optimal value for $k$ to achieve the best overall performance. Similarly, Fig.~\ref{fig:hyper}(d) shows that Dice and clDice also reach their peak at the same value. Including too many foreground pixels in the background window hinders the effective creation of \textit{pseudo breaks}, while overly strict constraints on the foreground pixel ratio result in an insufficient number of \textit{pseudo breaks}.

\vspace*{5pt}
\noindent\textbf{Impact of different synthesis qualities.}
Our background patch search strategy around neighbors, combined with low-frequency swap, ensures sample authenticity at most times. To further validate it, we conduct experiments with background patches of varying similarity, where higher similarity corresponds to better synthesis quality. 
Specifically, we compute the similarity between all patches in the image and the selected foreground patch. We then conduct low-frequency swap using the least similar, moderately similar, and most similar ones, respectively, and evaluate their individual performances.
As shown in Table~\ref{tab:sup5}, the performance of using neighbor patch (Ours) is almost identical to that of using the best quality patches (need extra time +3.02s per image), indicating the high quality of our synthesized samples. To trade off time and accuracy, our method remains the preferred choice. Note that even with the worst quality patches, performance still outperforms the second-best baseline.

\begin{table*}[h]
    \centering
	\caption{Ablation results of the different synthesis qualities. The best and second-best results in each column are highlighted in \textbf{bold} and \underline{underline}, respectively.}
        \vspace{-5pt}
	\setlength{\tabcolsep}{5pt}
        \renewcommand\arraystretch{1.1}
	\resizebox*{0.7 \linewidth}{!}{
\begin{tabular}{lcccccc}
            \toprule[1.2pt]
            \multirow{2}{*}{\textbf{Synthesis quality}} & \multicolumn{3}{c}{\textbf{DRIVE $\rightarrow$ CHASE}} & \multicolumn{3}{c}{\textbf{CHASE $\rightarrow$ DRIVE}}\\
    \cmidrule(lr){2-4}\cmidrule(lr){5-7}
     & Dice (\%) $\uparrow$ & clDice (\%) $\uparrow$ & $\beta$ $\downarrow$ & Dice (\%) $\uparrow$ & clDice (\%) $\uparrow$ & $\beta$ $\downarrow$ \\
            \midrule
            CoTTA & 68.60  & 71.53 & 36.38 & 67.64 & 64.80 & 81.20 \\
            \midrule
            Worst quality (least similar) & 69.02 & 73.82 & 26.13 & 71.75 & 67.65 & 84.05 \\
            Middle quality (moderately similar) & 70.69 & 75.88 & 25.87 & 72.55 & 68.85 & 82.00 \\
            Best quality (most similar) & \underline{70.23} & \textbf{77.65} & \textbf{23.35} & \textbf{73.22} & \textbf{70.46} & \underline{80.20} \\
            \textbf{Neighbor} & \textbf{70.73}  & \underline{77.05}  &  \underline{25.38} & \underline{72.96}  &  \underline{70.26} & \textbf{79.15} \\
            \bottomrule[1.2pt]
          \end{tabular}
	}
    \vspace*{-20pt}
    \label{tab:sup5}
\end{table*}

\vspace*{5pt}
\noindent\textbf{Necessity of updating router parameter.}
To verify that simultaneously adjusting both \textbf{$\delta$} and the model parameters increases the search-space complexity, we conduct an experiment where all parameters are updated together. As shown in Table~\ref{tab:sup6}, the results reveal a significant performance drop, indicating that this approach may introduce greater parameter instability. And in the paper's setting, router parameters \textbf{$\delta$} add only 1280 params, a negligible increase compared to the original model's params ($2.894\times10^6$), and fewer than methods like VPTTA, which add 4332 params.

\begin{table*}[h]
    \centering
	\caption{Ablation results of adjusting either all parameters or only router parameters \textbf{$\delta$} at Stage 1.}
        \vspace{-5pt}
	\setlength{\tabcolsep}{5pt}
        \renewcommand\arraystretch{1.1}
	\resizebox*{0.7 \linewidth}{!}{
          \begin{tabular}{cccccccc}
            \toprule[1.2pt]
            \multirow{1}{*}{\textbf{Upadate}} & \textbf{Param} & \multicolumn{3}{c}{\textbf{DRIVE $\rightarrow$ CHASE}} & \multicolumn{3}{c}{\textbf{CHASE $\rightarrow$ DRIVE}}\\
    \cmidrule(lr){3-5}\cmidrule(lr){6-8}
      \textbf{params} & \textbf{size} & Dice (\%) $\uparrow$ & clDice (\%) $\uparrow$ & $\beta$ $\downarrow$ & Dice (\%) $\uparrow$ & clDice (\%) $\uparrow$ & $\beta$ $\downarrow$ \\
            \midrule
            All & $2.895\times10^6$ & 70.07 & 72.58 & 30.23 & 69.08 & 65.77 & 81.7 \\
            \textbf{$\delta$} & 1280 & \textbf{70.73} & \textbf{77.05} & \textbf{25.38} & \textbf{72.96}  & \textbf{70.26} & \textbf{79.15} \\
            \bottomrule[1.2pt]
          \end{tabular}
	}
    \vspace*{-5pt}
    \label{tab:sup6}
\end{table*}

\begin{figure}[h]
	\setlength{\abovecaptionskip}{6pt}
	\setlength{\belowcaptionskip}{-5pt}
	\centering 
	\includegraphics[width=0.6\textwidth]{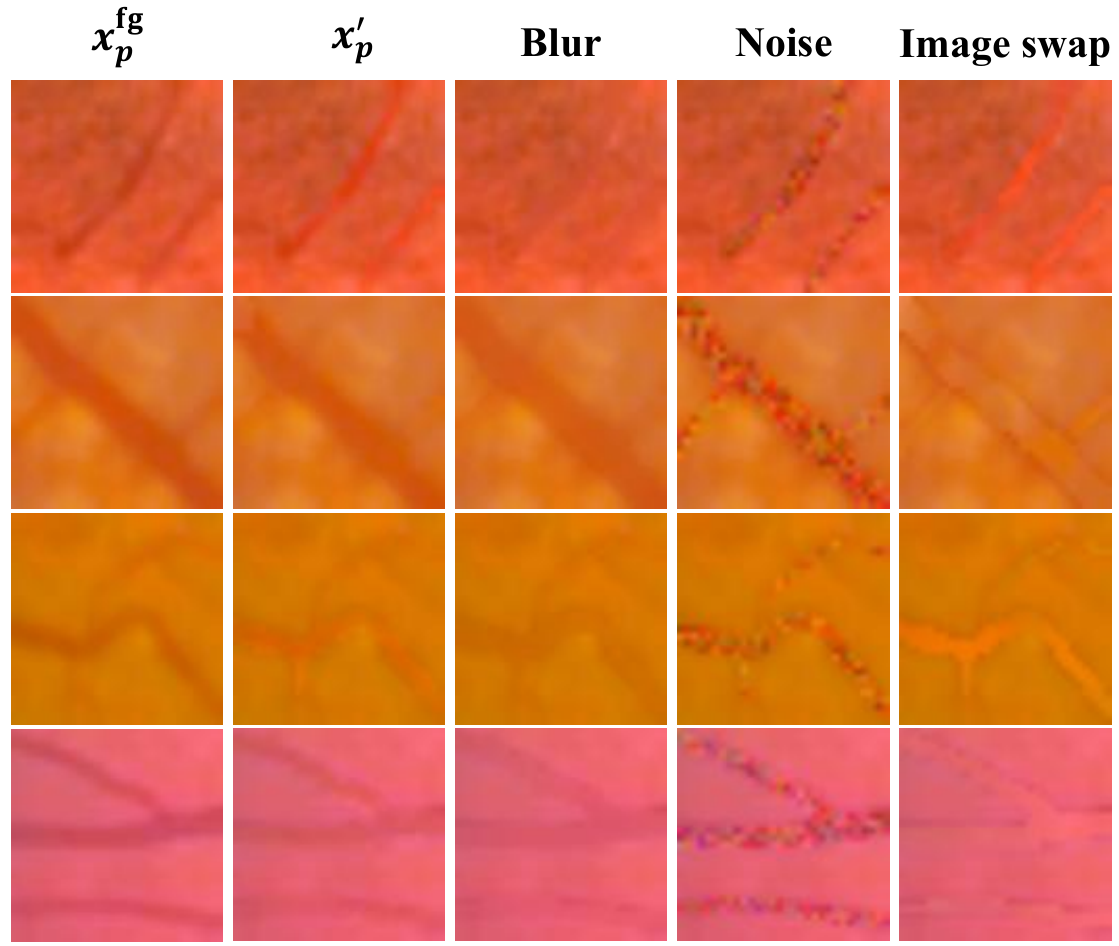}
	\caption{Visualizations of pseudo-breaks generated by TopoHG and three data augmentation methods, i.e., Gaussian blur, random Gaussian noise, and image swap in the spatial domain. $\boldsymbol x_p^{\text{fg}}$ is original patch and $\boldsymbol{x}_{p}^{\prime}$ denotes pseudo-break generated by TopoHG.} 
	\label{fig:dif_aug} 
\end{figure}

\vspace*{5pt}
\noindent\textbf{Visualization of Different Data Augmentation Methods.}
We visualize the effects of different data augmentation methods, as shown in Fig.~\ref{fig:dif_aug}. Our frequency-based method effectively preserves high-frequency details in the generated \textit{pseudo-breaks}. In comparison, the Gaussian blur method shows moderate visual effects, while random Gaussian noise and image swap methods overly modify the images, creating actual breaks.

\subsection{Additional Visualization Results}


\begin{figure*}[h]
	\centering 
	\includegraphics[width=1.0\textwidth]{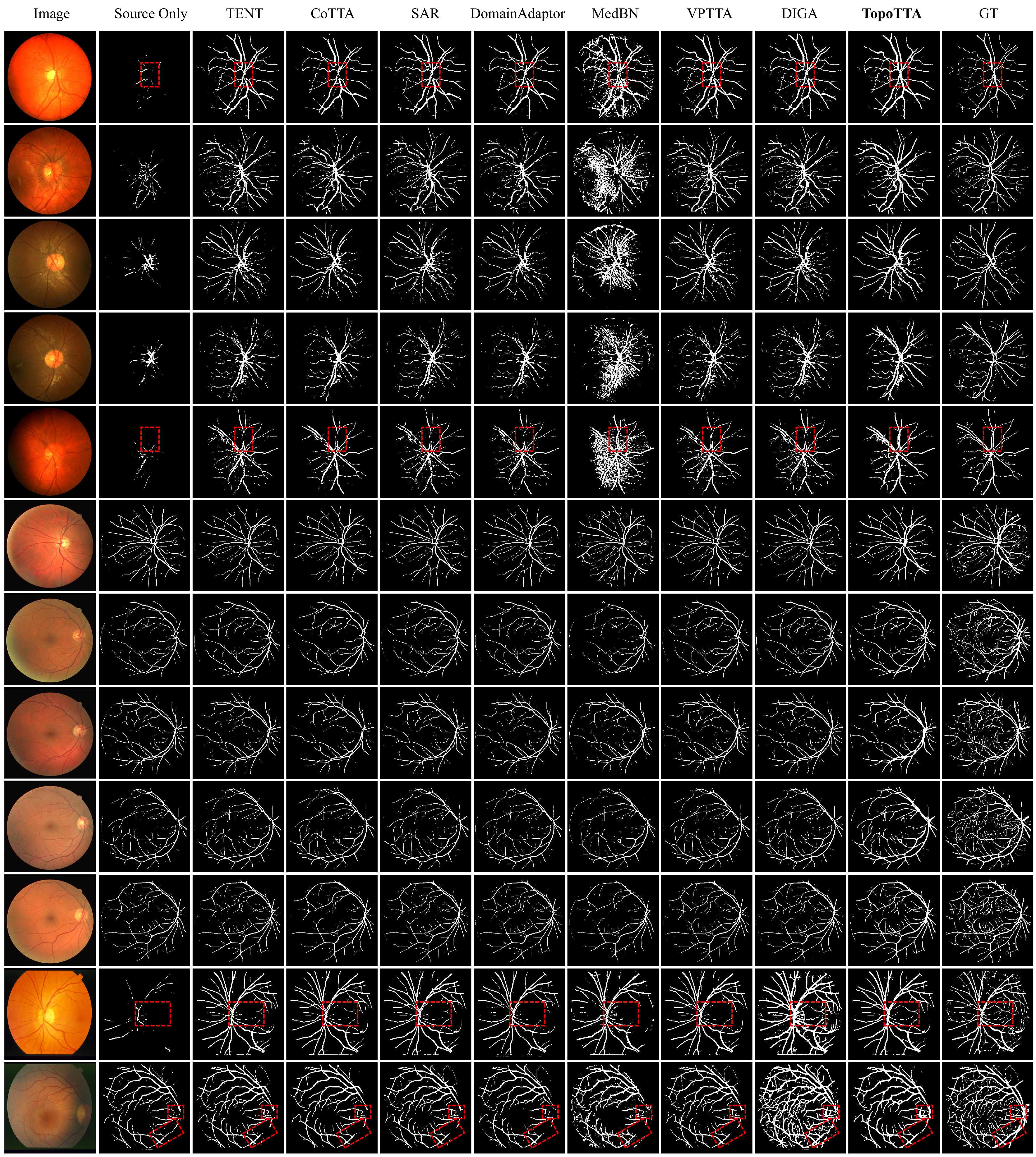}
	\caption{Visualization of segmentation results for TopoTTA and seven comparison methods in retinal vessel segmentation scenario. “Source Only” denotes results without any TTA methods applied, and GT is short for ground-truth labels.}  
\end{figure*}

\begin{figure*}[t]
	\centering 
	\includegraphics[width=1.0\textwidth]{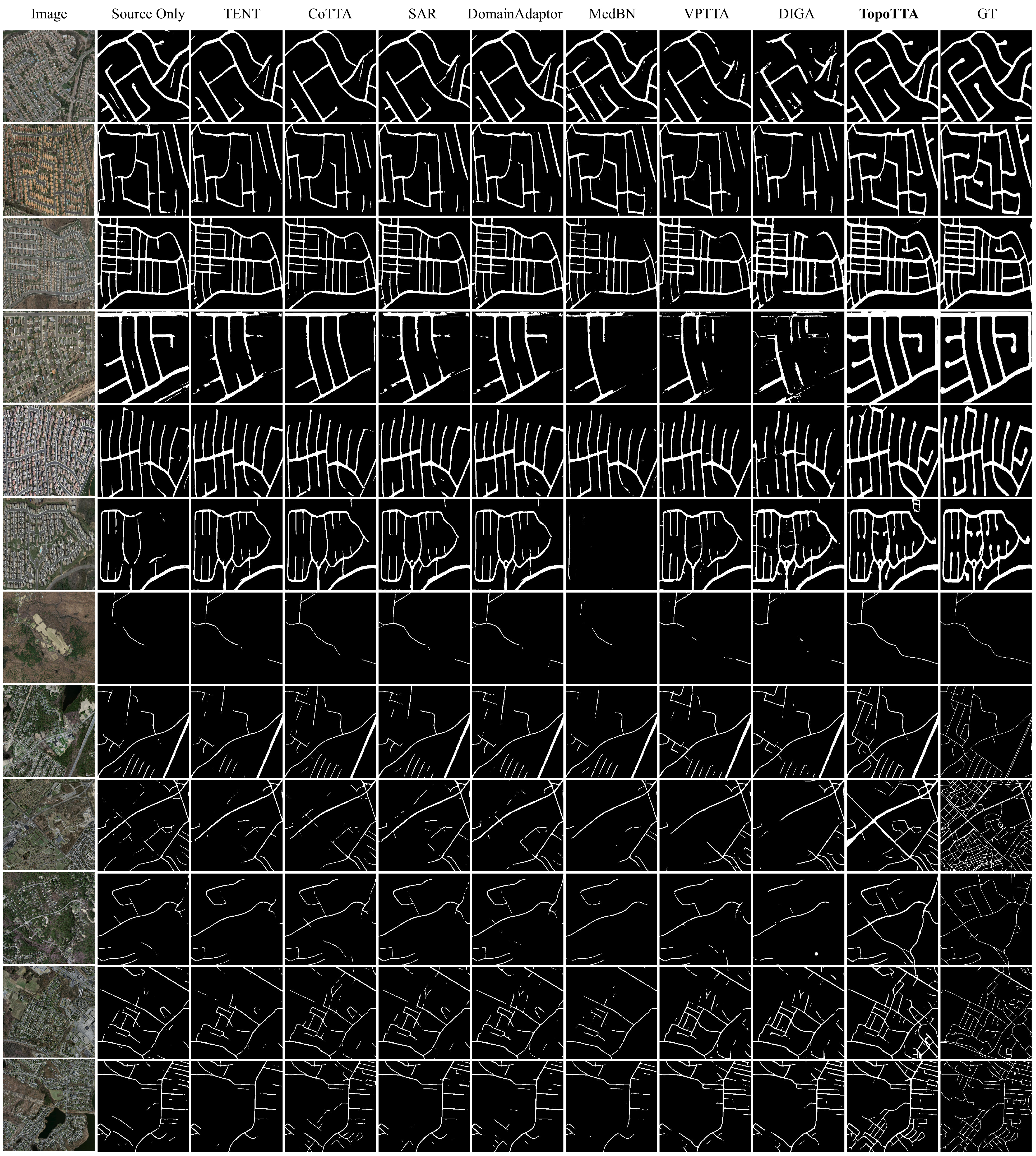}
	\caption{Visualization of segmentation results for TopoTTA and seven comparison methods in road extraction scenario. “Source Only” denotes results without any TTA methods applied, and GT is short for ground-truth labels.} 
\end{figure*}

\begin{figure*}[t]
	\centering 
	\includegraphics[width=1.0\textwidth]{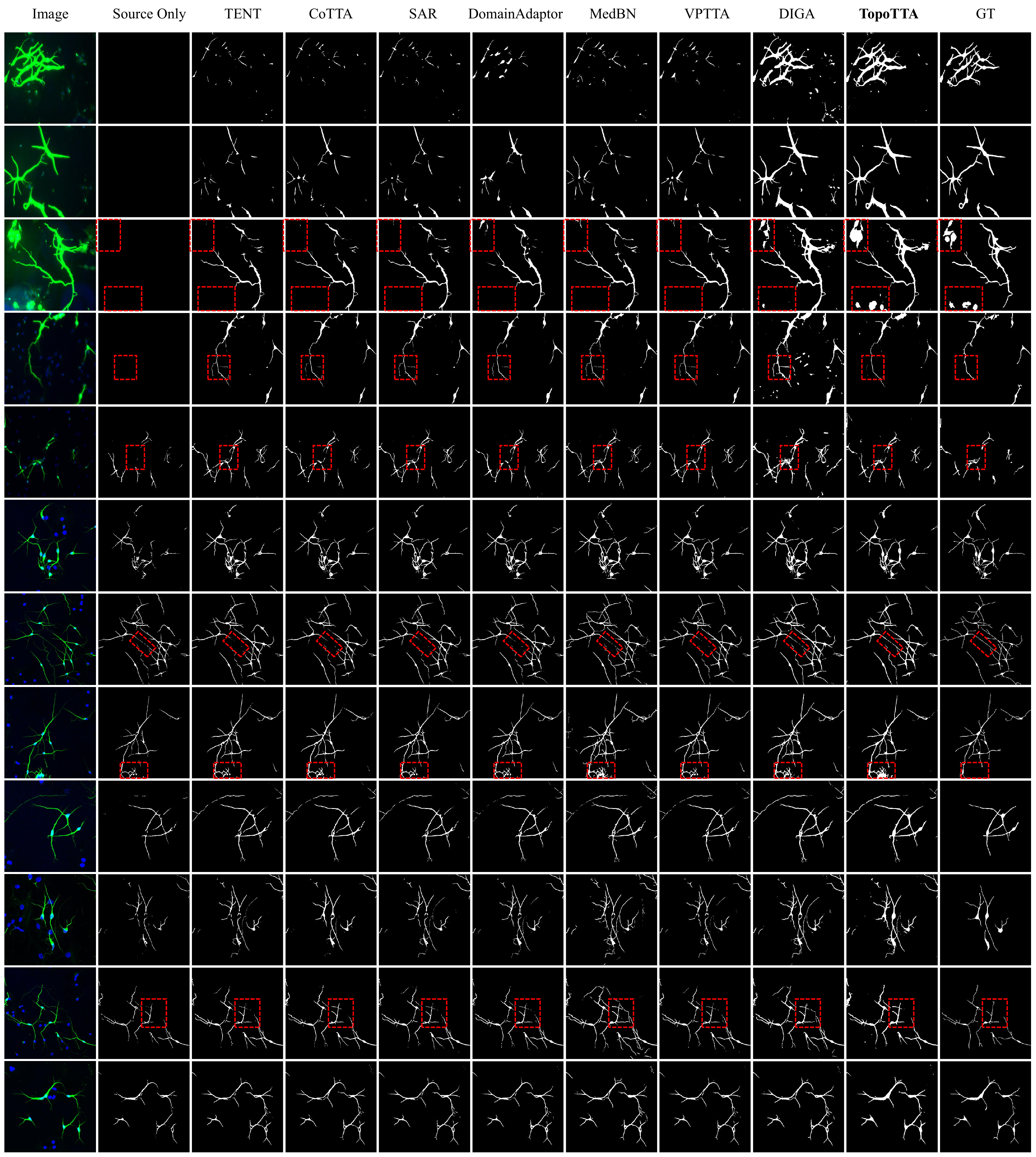}
	\caption{Visualization of segmentation results for TopoTTA and seven comparison methods in microscopic neuronal
segmentation scenario. “Source Only” denotes results without any TTA methods applied, and GT is short for ground-truth labels.} 
\end{figure*}

\begin{figure*}[t]
	\centering 
	\includegraphics[width=1.0\textwidth]{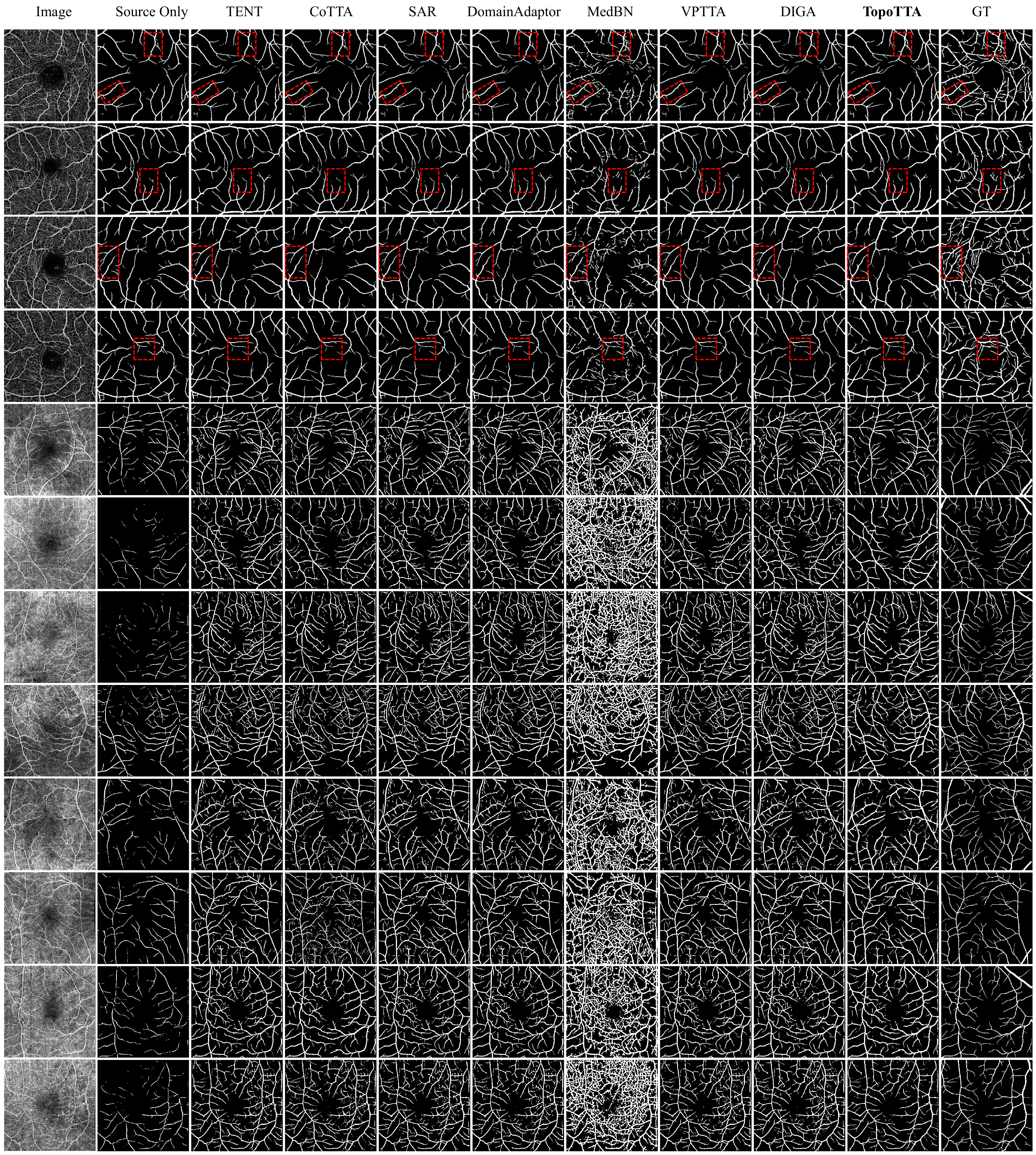}
	\caption{Visualization of segmentation results for TopoTTA and seven comparison methods in retinal OCT-angiography vessel segmentation scenario. “Source Only” denotes results without any TTA methods applied, and GT is short for ground-truth labels.} 
\end{figure*}